\pdfoutput=1

\documentclass[11pt]{article}

\usepackage{ACL2023}

\usepackage{times}
\usepackage{latexsym}

\usepackage[T1]{fontenc}

\usepackage[utf8]{inputenc}

\usepackage{microtype}

\usepackage{inconsolata}
\usepackage{amssymb}
\usepackage{amsmath}
\usepackage{amsthm}
\usepackage{graphicx}
\usepackage{mathtools}
\graphicspath{ {./img/} }

\usepackage{multirow}
\usepackage{hyperref}
\usepackage{booktabs}
\usepackage{enumitem}
\usepackage{algorithm2e}
\RestyleAlgo{ruled}

\setlist[itemize]{align=parleft,left=0pt..1em}

\makeatletter
\renewcommand{\maketag@@@}[1]{\hbox{\m@th\normalsize\normalfont#1}}%
\makeatother

\title{NLU on Data Diets:\\Dynamic Data Subset Selection for NLP Classification Tasks}

\author{Jean-Michel Attendu\thanks{\hspace{0.2cm}Equal contributions.} \and Jean-Philippe Corbeil$^*$ \\
  Nuance Communications \\
  \texttt{\{jattendu,jcorbeil\}@microsoft.com}
}

\begin{document}
\maketitle

\begin{abstract}
Finetuning large language models inflates the costs of NLU applications and remains the bottleneck of development cycles. Recent works in computer vision use data pruning to reduce training time. Pruned data selection with static methods is based on a score calculated for each training example prior to finetuning, which involves important computational overhead. Moreover, the score may not necessarily be representative of sample importance throughout the entire training duration. We propose to address these issues with a refined version of dynamic data pruning, a curriculum which periodically scores and discards unimportant examples during finetuning. Our method leverages an EL2N metric that we extend to the joint intent and slot classification task, and an initial finetuning phase on the full train set. Our results on the GLUE benchmark and four joint NLU datasets show a better time-accuracy trade-off compared to static methods. Our method preserves full accuracy while training on 50\% of the data points and reduces computational times by up to 41\%. If we tolerate instead a minor drop of accuracy of 1\%, we can prune 80\% of the training examples for a reduction in finetuning time reaching 66\%.
\end{abstract}

\section{Introduction}

\begin{figure}[ht!]
\includegraphics[scale=0.85]{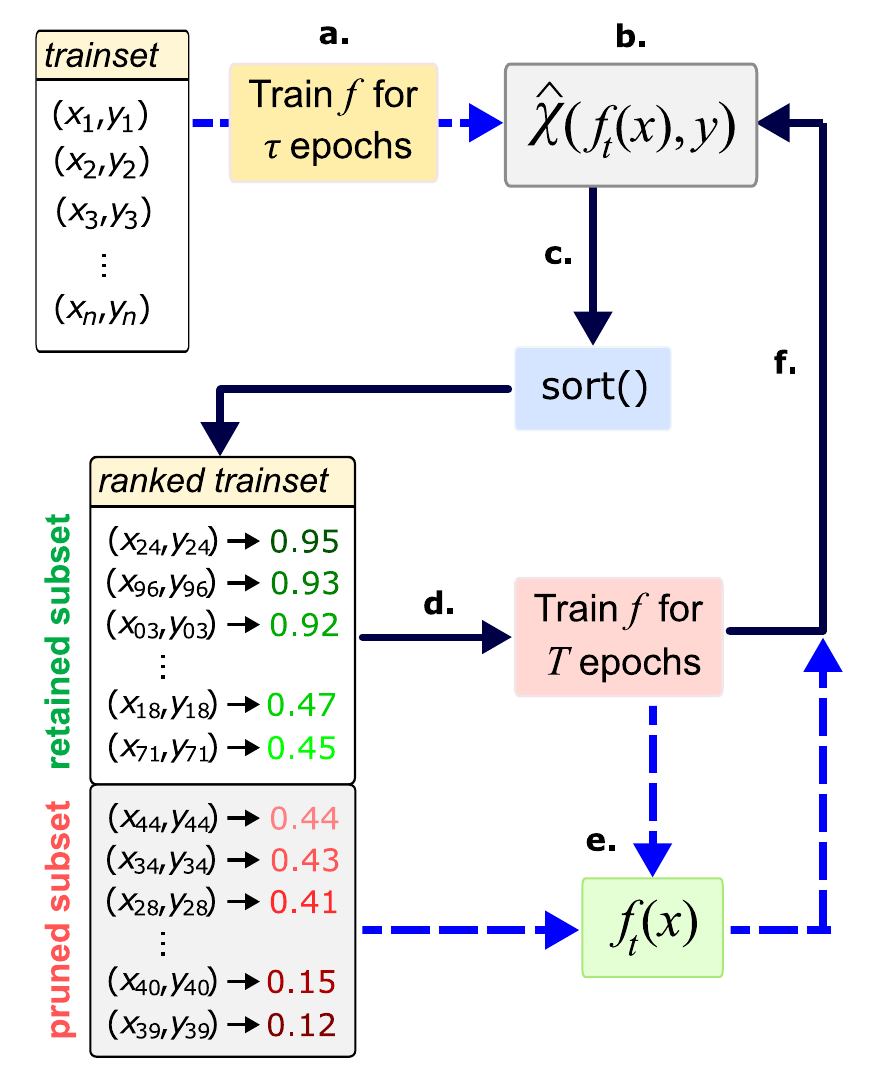}
\caption{\textit{a}.~Model $f$ is trained for $\tau$ epochs with full train set. \textit{b}.~Importance function $\hat{\chi}$ assigns score to each training example $(x_i,y_i)$. \textit{c}.~Training examples are sorted according to their scores. \textit{d}.~Proportion of train set $(1-\rho)$ is retained to train model over a cycle of $T$ epochs. \textit{e}.~Resulting model is used to calculate a forward pass. \textit{f}.~Full training set is re-scored for the next cycle until completion.}
\label{fig:method_diagram}
\end{figure}

State-of-the-art natural language understanding (NLU) systems rely on finetuning large language models (LLMs) with considerable amounts of human-labelled examples. Such overparametrized models succeed at learning to classify nearly all training samples correctly. However, not all training examples are created equal: proper classification of out-of-distribution outliers requires more iterations for the model to converge. Moreover, many examples are correctly learned after only a few epochs of training and could be ignored further on without impacting test accuracy.

This raises a question about data efficiency: can we remove a proportion of training examples to reduce the number of calculations during finetuning while preserving accuracy at inference? Since finetuning LLMs is computationally intensive and represents a bottleneck in time and costs of development cycles of NLU applications, such an approach appears promising.

Previous works in computer vision suggest that an important fraction of training examples, up to 50\% for some datasets, can be removed without significant loss in accuracy \cite{toneva2018empirical,paul2021deep,raju2021accelerating}. Assuming total train steps increase proportionally with the number of samples, this represents a considerable reduction of training costs while being completely compatible with other existing strategies to make finetuning more efficient \cite{gupta2015deep,micikevicius2017mixed,zhang2020accelerating,zaheer2020big,xiong2021nystromformer}. The core idea is to score each training example $(x_i,y_i)$ with an importance metric, and to remove a proportion of the samples based on ranked scores: the most important examples are kept, whereas others are discarded.


In this work, we expand the analysis of such a method to text classification across various datasets. In agreement with \citet{raju2021accelerating}, we observe that many examples have low importance when evaluated at early training but with high variability further on. Thus, a single evaluation often fails at predicting their impact on the loss at a later stage. To address this problem, we also adopt a curriculum in which pruning is performed periodically at a fixed number of epochs. The model is first trained with all examples for a few epochs before ranking is performed with respect to metric $\hat{\chi}$. Then, a proportion $\rho$ of the samples with the lowest scores is set aside, and the model is trained with the retained subset for a cycle of $T$ epochs. All examples are re-scored at every cycle's end to obtain an updated subset. New cycles are computed until training is complete. Since $\hat{\chi}$ is cheap to compute, the overhead of periodic calculations for a new pruned subset remains low. Figure \ref{fig:method_diagram} provides an overview of our proposed method.

Our contributions are as follows:

\begin{itemize}
    \item  We investigate the effectiveness of data pruning methods applied to text classification in reducing finetuning time while maintaining accuracy, including experimental results on six datasets from the GLUE benchmark and four joint NLU datasets.
    \item We improve the dynamic pruning method with a key initial training phase on the full train set and novel application of the EL2N score in a dynamic setting.
    \item We extend the EL2N importance score to multi-output text classifications supported by mathematical derivations, specifically for joint intent classification and slot filling.
    \item We release an open source version of our implementation\footnote{\url{github.com/jpcorbeil-nuance/nlu\_data\_diets}} built around the \textit{transformers} library \cite{wolf2020transformers}.
\end{itemize}

\section{Related Work}

Previous works have investigated data pruning methods based on metrics calculated from training characteristics. \citet{toneva2018empirical} propose to score examples based on their number of transitions from correct to incorrect classification over the course of training. They observed that training accuracy is not considerably affected by removing a significant proportion of the rarely forgotten training examples. Nevertheless, this method is not directly applicable to efficiency improvements since evaluating forgetting scores requires a full round of finetuning. \citet{coleman2019selection} address this issue by conducting data selection using a small proxy model. However, this involves supporting an additional model, which makes the process more complex. \citet{paul2021deep} derives GraNd and EL2N scores as approximations to sample importance \cite{alain2015variance,katharopoulos2018not}. They observe that expectation over 10 initializations improves metric quality when evaluated early in training. Although this represents a considerable computational cost, they show that they can prune large proportions of low-importance samples without test accuracy reductions. Others have applied GraNd/EL2N scores to text datasets, notably in natural language inference \cite{fayyaz2022bert} and to reduce gender bias \cite{zayed2022deep}. \citet{sorscher2022beyond} criticize previous work for experimenting only on two computer vision datasets: CIFAR-10 and CIFAR-100. Other metrics have been proposed \cite{sorscher2022beyond,mindermann2022prioritized,yang2022influence}, but they all involve further computations.


\citet{raju2021accelerating} and \citet{swayamdipta2020dataset} observe that training dynamically separates examples in three categories: easy-to-learn, hard-to-learn and ambiguous. Ambiguous samples are characterized by higher variability across training, and an early evaluation of their importance metric can incorrectly predict late training behaviour. To address this, a curriculum is proposed to train on dynamically selected subsets, leading to better performance than EL2N-based static pruning and random subset selection at high pruning rates.

Other related topics also include coreset selection \cite{har2005smaller,munteanu2018coresets,campbell2018bayesian}, active learning \cite{ash2019deep,kim2020deep,park2022active}, dataset size reduction \cite{mishra2020we} and curriculum learning \cite{bengio2009curriculum,kumar2010self,platanios2019competence,wan2020self,christopoulou2022training}.

\section{Methods}

\subsection{Estimating Training Example Importance}

Importance refers to the degree to which a particular training example affects the model's performance at learning to correctly classify other examples, as formally defined in \citet{paul2021deep}. A low-importance sample has little impact on training and can be removed without much effect.

We consider supervised classification, with training set $S$ composed of training examples $(x_i,y_i)$ with $i \in \left[1,N\right]$ and $N \in \mathbb{N}$, drawn i.i.d. from an unknown data distribution. Here, $x_i \in \mathbb{R}^d$ are the input vectors of dimension $d \in \mathbb{N}$ and $y_i \in \{0,1\}^K$ are one-hot labels with $K$ classes. Model $f$, composed of weights $w \in \mathcal{W} \subseteq \mathbb{R}^D$ with $D\in\mathbb{N}$, produces probability vector $p=f(x_i) \in \left[0,1\right]^K$, and is trained to estimate $y_i$ using the cross-entropy loss:
\begin{equation}
\mathcal{L}_i = -y_i^T \cdot log(f(x_i)).
\label{eq_celoss}
\end{equation}

\citet{paul2021deep} showed that the norm of the gradient of $\mathcal{L}_i$ is an upper bound to the importance of sample $(x_i,y_i)$. They also introduced a computationally more efficient approximation known as the EL2N score: 

\begin{equation} 
\chi(x_i,y_i) = {\mathbb{E}}||f(x_i) - y_i||_2, 
\end{equation}

where the expectation is taken over the randomly initialized weights $w$ of model $f$.

\subsubsection{Extending EL2N to NLU}
For the joint NLU task, each training example has sequence-level intent labels $y_i^{intent}$ and $M \in \mathbb{N}$ word-level slot labels $y_{i,m}^{slot}$ with $m \in \{1,M\}$. The loss for sample $i$ is expressed as:
\begin{equation}
\mathcal{L}_i^{nlu} = \mathcal{L}_i^{intent} + \sum_{m=1}^{M}\mathcal{L}_{i,m}^{slot},
\label{eq_nluloss}
\end{equation}
\noindent where intent and slot losses are calculated from eq.~\ref{eq_celoss} with $y_i^{intent}$ and $y_{i,m}^{slot}$ respectively.

In order to estimate joint NLU sample importance, we extend the EL2N formulation to a such case in Appendix \ref{sec:el2n_equations}. We provide a derivation for the EL2N score in three contexts: EL2N for sequence-level classification:
\begin{equation}
    \hat{\chi}_{intent}(x_i,y_i) = ||f(x_i) - y_i^{intent}||_2,
\end{equation}
\noindent EL2N for multiple-output classifications, which introduces a second $\ell_2$ norm over the sequence length:
\begin{equation}
    \hat{\chi}_{slot}(x_i,y_i) = \sqrt{\sum_{m=1}^{M}||f(x_i) - y_{i,m}^{slot}||_2^2},
\label{eq_slots}
\end{equation}
\noindent and for joint tasks, e.g. joint intent detection and slot filling:
\begin{equation}
   \hat{\chi}_{nlu}(x_i,y_i) = \sqrt{\hat{\chi}_{intent}(x_i,y_i)^2  + \hat{\chi}_{slot}(x_i,y_i)^2 }.
\end{equation}

\subsubsection{Averaging $\hat{\chi}$ Over Time}
Averaging EL2N over multiple initializations improves importance approximation, but increases compute costs. Instead, following \citet{raju2021accelerating}, we consider an exponential moving average (EMA) over time to evaluate $\hat{\chi}$:
\begin{equation}
  \begin{aligned}
     \hat{\chi}_{ema}(x,y) \leftarrow &\alpha \cdot \hat{\chi}_{nlu}(x,y) \\ + & (1-\alpha) \cdot \hat{\chi}_{ema}(x,y)
  \end{aligned}
\label{eq_ema}
\end{equation}
This is possible since dynamic pruning is performed periodically over multiple cycles. Because importance estimation for some examples varies significantly across train steps, EMA provides an adequate trade-off between emphasizing the current estimate and leveraging previous evaluations.

\subsection{Dynamic Pruning}
With dynamic pruning, the model is trained for $\tau$ epochs, then, we train for cycles of $T$ epochs until training is complete. At the beginning of each cycle, we score all training examples with eq. \ref{eq_ema}, and the proportion with the highest scores is selected. In algorithm \ref{alg:dynprune}, we present pseudo-code for the proposed method, where $S'$ is the retained training subset and $E$ is the number of training epochs.

\SetKwInput{KwData}{Inputs}
\SetKwComment{Comment}{/* }{ */}
\begin{algorithm}
\caption{Dynamic Pruning}\label{alg:dynprune}
\KwData{$S$, $f$, $\rho$, $T$, $\tau$, $E$}
$S'\gets \{\}$\;
$\hat{\chi}_{ema}\gets \{\}$\;
$cycle \gets 1$\;
\For{epoch in $\tau$}
    {
       train $f$ with $S$\;
    }
\While{$cycle < \lfloor($E$-\tau)/T \rfloor$}
{
    $\hat{\chi}_{ema}  \gets \alpha\hat{\chi}_{nlu}+(1-\alpha)\hat{\chi}_{ema}$\;
    \emph{sort} $S$ with respect to $\hat{\chi}_{ema}$, $\downarrow$ order\;
    $S' \gets$ top $(1-\rho)$ elements from $S$\;
    $cycle \gets cycle+1$\;
{
    \For{epoch in $T$}
    {
       train $f$ with $S'$\;
    }
}}
\end{algorithm}

\section{Experimental Details}
\subsection{Experimental Setup}

In our experiments, we compare five methods: 

\begin{itemize}
    \item \textbf{Full Train Set}: baseline approach with standard training using the full train set.
    \item \textbf{Static Pruning}: method from \citet{paul2021deep}. The training subset is obtained from EL2N scores averaged over 10 randomly initialized models trained for 10 epochs, and finetuning is then performed on the fixed subset.
    \item \textbf{Single Pruning}: training is performed on the full trainset for $\tau$ epochs, followed by a single subset selection using EL2N. Finetuning is then performed on the fixed subset.
    \item \textbf{Dynamic Pruning}: proposed method which trains on the full trainset for $\tau$ epochs, followed by periodic data pruning using $\hat{\chi}_{ema}$ as described by algorithm \ref{alg:dynprune}.
    \item \textbf{Dynamic Random Pruning}: same as dynamic pruning, but the subset selection is random.
\end{itemize}

For the full train set method, we finetune the model for $E$ epochs. We assume that $E$ and $\tau$ correspond to constant numbers of training steps defined from the full training set size. For the other four methods, all experiments are performed over $E$ epochs, to which we remove a number of train steps equivalent to $\rho \cdot (E-\tau)$. We do not report results for the full train set method with reduced training steps as they are comparable to those of the dynamic random pruning method. The number of epochs $E$ is set to 10 for GLUE following \citet{liu2019roberta} and to 40 for the joint NLU tasks following \citet{chen2019bert}.

\subsection{Datasets}
We present two series of experiments to address text classification problems with either a single output or multiple outputs. For the single output case, we consider six datasets from the GLUE benchmark \cite{wang2019glue}: COLA, MNLI, MRPC, QQP, RTE, and SST2. We exclude STS-B since it is a regression task, WNLI because it has a very small train set, and QNLI as it is an artificially built dataset.


For the multi-output case, we use four open-source joint NLU datasets: ATIS \cite{price1990atis,tur2010left}, SNIPS \cite{coucke2018snips}, SLURP \cite{bastianelli2020slurp} and MTOP \cite{li2021mtop} --- English version only. All examples from these datasets are labelled with an intent (sequence-level class) and entity tags (token-level class). ATIS focuses on the airline travel domain, and the other three datasets represent voice assistants across various domains, such as music, weather, and calendar, among others. We use the original versions of these datasets without any modification.

Tables \ref{tab:data_glue_stats} and \ref{tab:data_nlu_stats} in Appendix \ref{sec:data_stat} provide the general characteristics of these datasets. Overall, they cover a large diversity of train set sizes and tasks. For the joint NLU tasks, we have a variety of domains, intents and entities.

\begin{table*}[ht!]
    \centering
    \caption{Median performances on 5 runs for the \textit{dev} set of the GLUE benchmark for pruning rate of 50\%. Static pruning is done with scores averaged on 10 runs of 10 epochs over 10 GPUs. All datasets are using $\tau=1$ and $T=2$, except for RTE set to $\tau=3$ and $T=1$ (marked by the asterisk *).}
    \small
    \begin{tabular}{lcccccccc}
    \toprule
     & COLA & MNLI & MRPC & QQP & RTE & SST2 \\
    \midrule
    \textbf{Full Train Set} & & & & & & \\
    \hspace{0.3cm} Score & 0.600 & 0.864 / 0.862 & 0.892 / 0.923 & 0.911 / 0.880 & 0.751 & 0.932 \\
    \hspace{0.3cm} Time (s) &  158.8 &  10053.0 &  89.0 &  8214.5 &  107.5 &  1314.8 \\
    \midrule
    \textbf{Static Pruning} & & & & & & \\
     \hspace{0.3cm} Score & 0.587 & 0.852 / 0.850 & 0.857 / 0.892 & 0.895 / 0.864 & 0.517 & 0.933 \\
     \hspace{0.3cm} Relative Time & 152.6\% & 151.2\% & 150.7\% & 146.5\% & 150.8\% & 150.7\% \\
    \textbf{Single Pruning} & & & & & & \\
     \hspace{0.3cm} Score & 0.586 & 0.778 / 0.775 & 0.854 / 0.887 & 0.909 / 0.879 & 0.746* & 0.937 \\
     \hspace{0.3cm} Relative Time & 58.0\% & 56.5\% & 56.6\% & 53.1\% & 68.2\%* & 55.7\% \\
    \textbf{Dynamic Random Pruning} & & & & & & \\
     \hspace{0.3cm} Score & 0.578 & 0.853 / 0.855 & 0.863 / 0.902 & 0.899 / 0.867 & 0.753* & \textbf{0.938} \\
     \hspace{0.3cm} Relative Time & 56.0\% & 54.5\% & 55.1\% & 54.3\% & 65.8\%* & 52.6\% \\
    \textbf{Dynamic Pruning (ours)} & & & & & & \\
     \hspace{0.3cm} Score & \textbf{0.590} & \textbf{0.863} / \textbf{0.863} & \textbf{0.882} / \textbf{0.915} & \textbf{0.912} / \textbf{0.881} & \textbf{0.773}* &    0.932 \\
     \hspace{0.3cm} Relative Time &  64.4\% &  64.0\% &  66.5\% &  60.4\% &  77.2\%* &  62.0\% \\
    \bottomrule
    \end{tabular}
    \label{tab:glue_results}
\end{table*}

\subsection{Model Architecture}
We use the pre-trained transformer model \textit{RoBERTa base} \cite{liu2019roberta}. For the GLUE datasets, we use single sequences or concatenations of two sequences as model inputs for paraphrase detection and natural language inference (NLI) tasks. For the joint NLU tasks, we use the JointBERT approach \cite{chen2019bert}. To handle entities, we apply a mask on special tokens as well as subsequent subwords if a word is decomposed into multiple subword tokens. Additionally, we reinitialize the last layer of the encoder \cite{kovaleva2019revealing,tamkin2020investigating}, which can improve the final performance.

\subsection{Accuracy Evaluation}
We evaluate the performance of our models on the GLUE datasets following the original instructions. Since the test sets are not publicly available, we report the averaged results on the development sets following previous approaches from \cite{devlin2019bert,liu2019roberta}. Specifically, we report the accuracy metric for RTE and SST2, accuracy on the matched and mismatched test set splits for MNLI, and both accuracy and F-1 metrics for MRPC and QQP. For COLA, we compute the Matthews correlation coefficient. For the joint NLU tasks, we focus on full-sequence accuracy, which requires matching both the intent and all the entities. We also provide the intent accuracy and entity micro F1 in Appendix \ref{sec:further_results}.

\subsection{Runtime Evaluation}

We report GPU finetuning time in minutes or seconds. To avoid the impact of CPU overhead, such as pre-processing and batching, we measure the time for each training step and forward pass for scoring EL2N separately, and then sum the results at the end of the fine-tuning process. We provide the mathematical expressions for the total GPU runtime and the minimum cycle $T_{min}$ to improve efficiency in Appendix \ref{sec:time_proof}. Finally, we used 10 GPUs to parallelize the computation of EL2N scores with static pruning, as this method requires training 10 independent models.

\section{Results}
\subsection{GLUE Benchmark}

\begin{figure*}[ht!]
\includegraphics[width=0.95\linewidth]{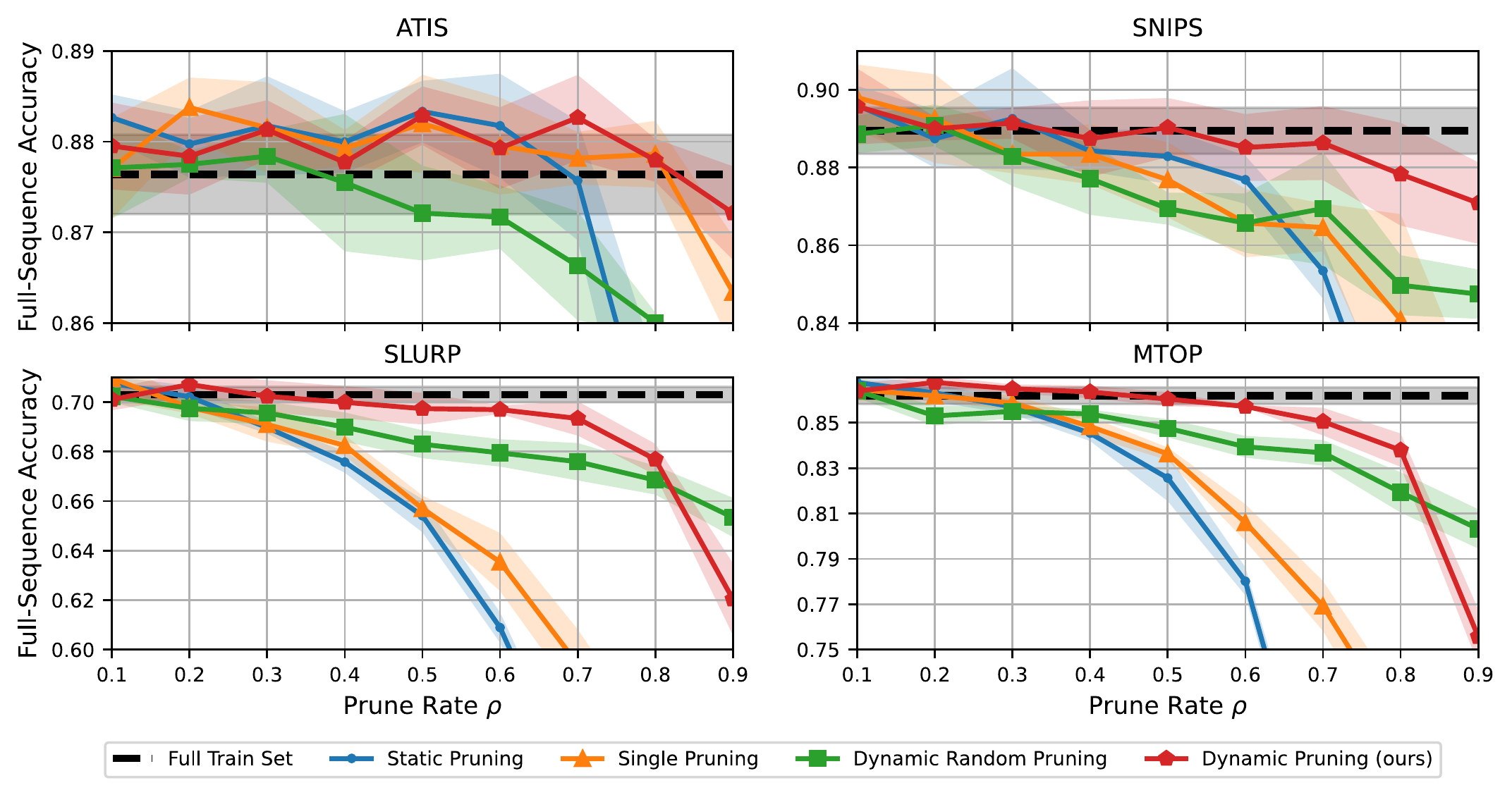}
\caption{Full-sequence accuracy achieved on 40 epochs for different prune rates on joint NLU datasets applying: static pruning (EL2N from 10 runs of 10 epochs), single pruning, dynamic random pruning, and our dynamic pruning (EL2N with EMA). The dynamic methods are run with $\tau=4$ and $T=4$.}
\label{fig:result_pruning_fullseq}
\end{figure*}

\begin{figure}[ht!]
\includegraphics[width=\linewidth]{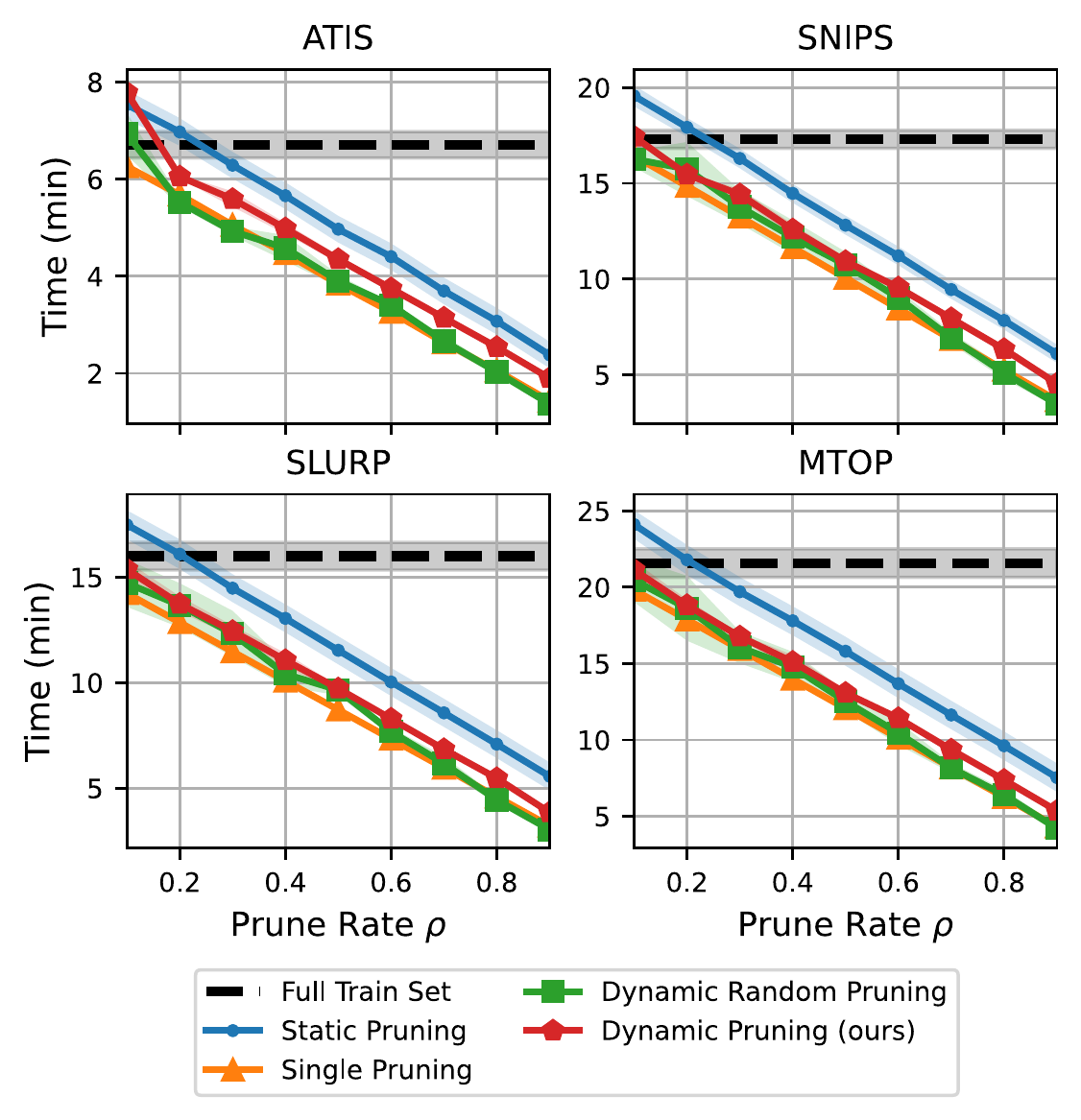}
\caption{Finetuning time  achieved on 40 epochs for different prune rates on joint NLU datasets  applying: static pruning (EL2N from 10 runs of 10 epochs), single pruning, dynamic random pruning, and our dynamic pruning (EL2N with EMA). The dynamic methods are run with $\tau=4$ and $T=4$.}
\label{fig:result_pruning_time}
\end{figure}

In Table \ref{tab:glue_results}, we present the results obtained on the GLUE benchmark with a 50\% pruning rate. Overall, dynamic pruning achieves accuracy nearly on par with full training while reducing the finetuning time by approximately 30\%. The largest accuracy decrease is observed on the COLA and MRPC datasets, at only 1\% absolute. We also note that the smallest time reduction occurs on RTE and is due to the different pruning settings required by this dataset of smaller size. We observe a sharper drop in performance with single pruning and dynamic random pruning, despite being faster. Static pruning underperforms in terms of both accuracy and finetuning time. This is especially the case on the RTE dataset and is likely due to the combination of a high pruning rate and a smaller number of training examples. This suggests that dynamic pruning may be affected at lower data amounts, as reported for static pruning by \citet{sorscher2022beyond}.

\subsection{Joint Intent Classification and Slot Filling}
\label{sec:result_nlu}

In Figures \ref{fig:result_pruning_fullseq} and \ref{fig:result_pruning_time}, we present the full-sequence accuracy and GPU finetuning time obtained with the studied methods on the joint NLU tasks. In Appendix \ref{sec:further_results}, we also include the related intent accuracy and slot micro F1 score.

\subsubsection{Accuracy Analysis}
Overall, our dynamic pruning method achieves accuracy comparable to full training, even when using aggressive pruning rates. It exhibits a flatter and smoother decrease compared to other techniques, which results in a better time-accuracy trade-off. For instance, at a pruning rate of 40\%, our approach leads to similar scores as the full-training baseline while being about 25\% faster. Other methods underperform by 0.5 to 2.5\% in the same setting, excluding the ATIS dataset for which all methods perform above baseline. Single pruning outperforms static pruning in almost every setting, despite static pruning leveraging EL2N scores calculated from multiple models. This emphasizes the importance of the initial training phase on the full dataset. Surprisingly, dynamic random pruning performs better than static and single pruning methods. This is further discussed in section \ref{sec:select_analysis}.

At an aggressive pruning rate of 80\%, dynamic pruning only incurs a marginal drop of 1-2\%, with a total finetuning time of nearly a third of the full training. In comparison, dynamic random pruning achieves a few absolute percentage points less than dynamic pruning, and static pruning decreases even faster. Single pruning shows a similar decrease, except on the ATIS dataset. At an extreme prune rate of 90\%, dynamic random pruning performs better than dynamic pruning for two datasets: SLURP and MTOP. This is because these datasets have many mislabeled data points, usually associated with high-importance scores, as discussed in section \ref{sec:select_analysis}. We also notice a significant decrease in accuracy from the static and single pruning methods for the same reason. Regarding domains, number of intents, and number of entities, SLURP and MTOP are the most complex.

Results on the ATIS dataset show a different trend, especially at a low prune rate. We observe an improvement over the full training set method for single, static and dynamic pruning, which perform equally well. ATIS is a smaller dataset with simple, repetitive samples on a single domain and consequently contains a larger proportion of easy-to-learn examples (see Section \ref{sec:select_analysis}). Even at moderate prune rates, the composition of the pruned subset remains stable over time, explaining the adequate performance of the static method. At higher prune rates, the accuracy of static pruning drops abruptly.



Figures \ref{fig:result_pruning_intent} and \ref{fig:result_pruning_slot} of Appendix \ref{sec:further_results} show that our dynamic pruning method preserves the full-trainset intent accuracy and slot micro F1 score with prune rates up to 80\%.

\subsubsection{Finetuning Time Analysis}
Regarding finetuning time, we empirically observe the linearly decreasing relationship with $\rho$ as described by equation \ref{eq:time_rho} in Appendix \ref{sec:time_proof}. Dynamic random pruning is faster than dynamic pruning since it does not require the periodic forward passes needed for calculating $\hat{\chi}_{ema}$. However, we can observe from this difference that the overhead is relatively small compared to the whole training process. Single pruning, which only requires a single computation of the importance score, is faster than dynamic pruning. In contrast, static pruning is significantly slower due to the overhead from the initial 10 runs of 10 epochs, even if computations are parallelized across ten GPUs.

\subsection{Balancing Accuracy and Efficiency}

\begin{figure}[ht!]
\includegraphics[width=\linewidth]{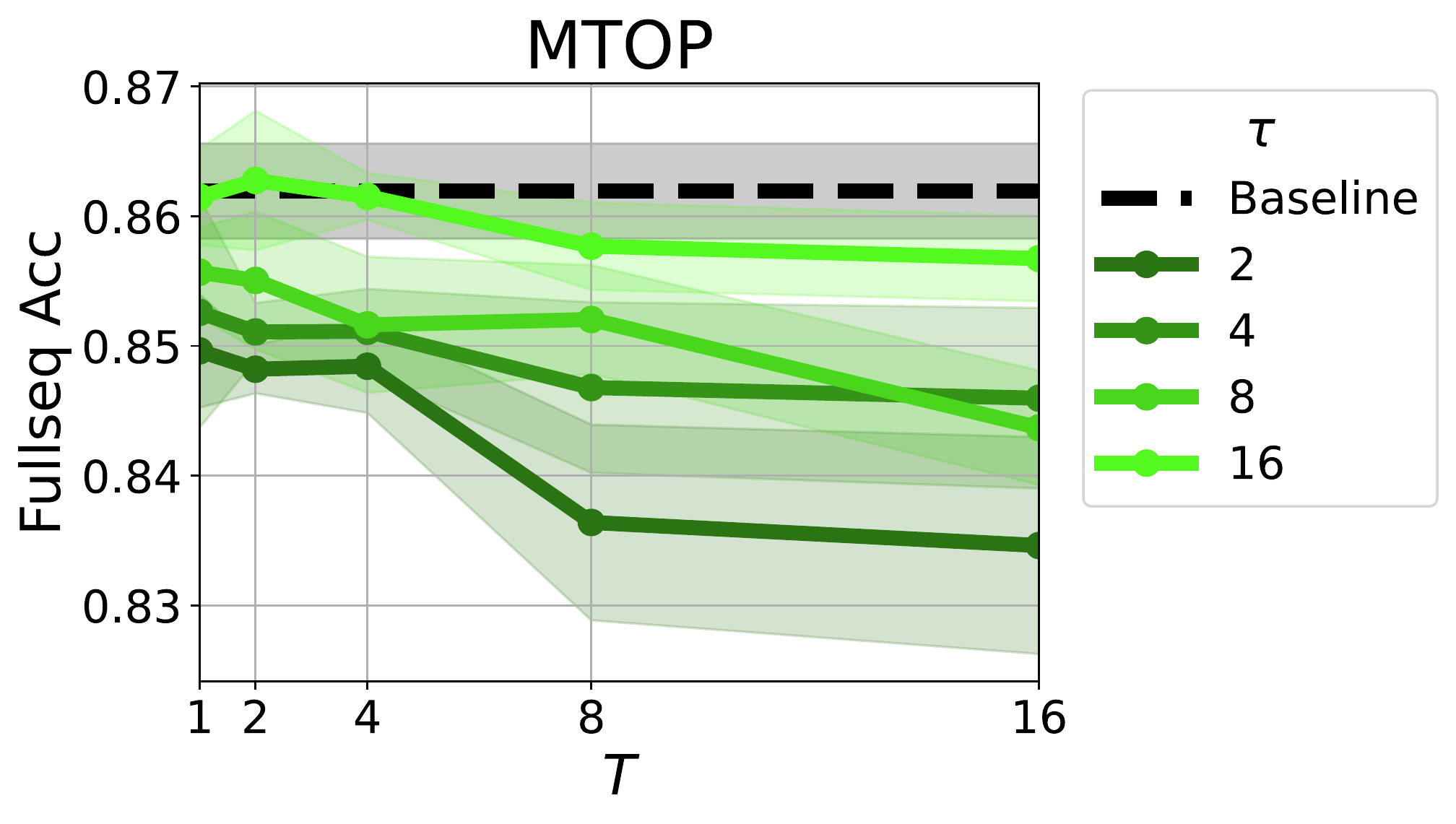}
\caption{Full-sequence accuracy from a dynamic pruning of 70\% for various $\tau$ and $T$ on the MTOP dataset.}
\label{fig:result_freq_mtop}
\end{figure}

\begin{figure}[ht!]
\includegraphics[width=0.97\linewidth]{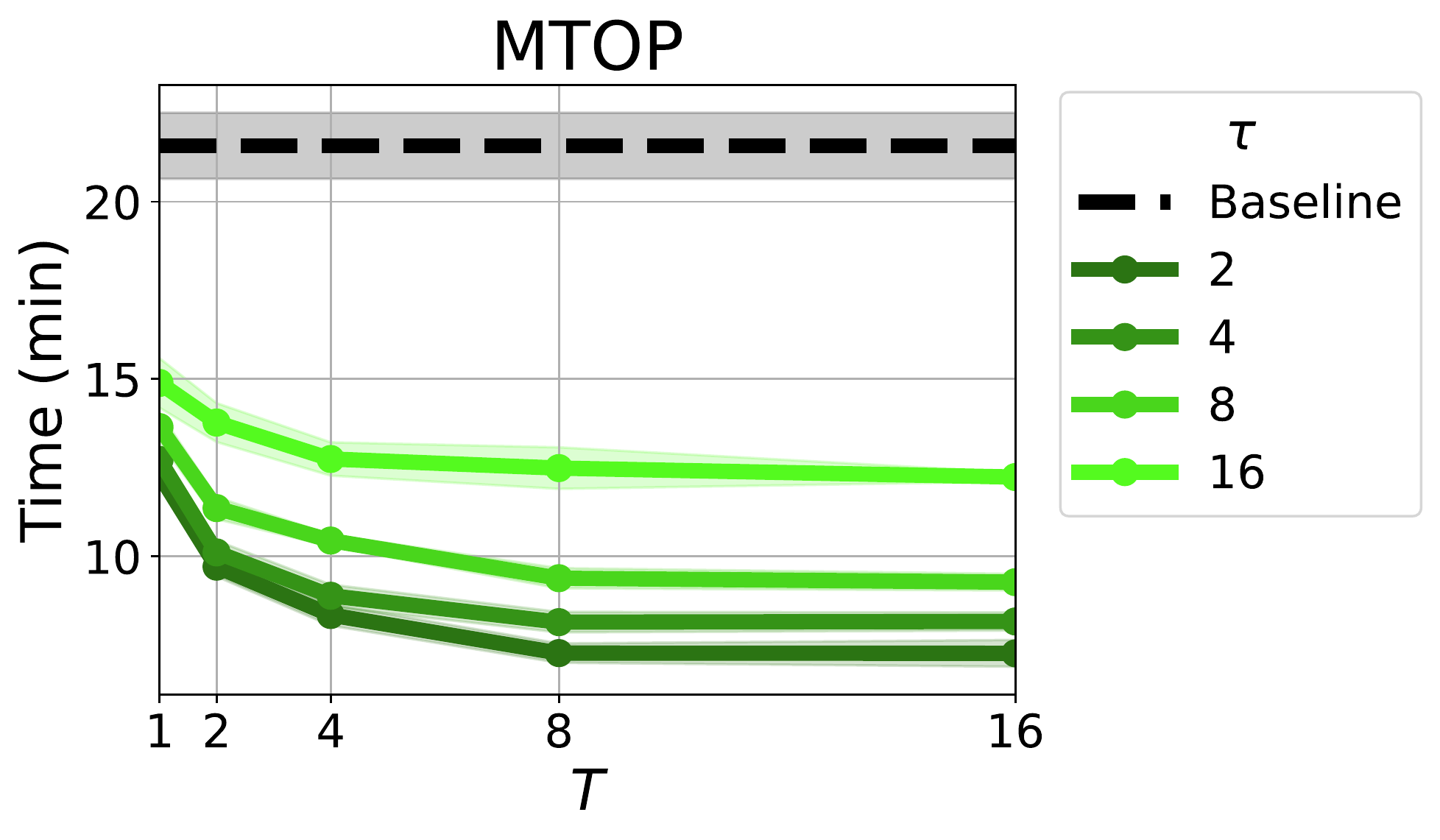}
\caption{Finetuning time from a dynamic pruning of 70\% for various $\tau$ and $T$ on the MTOP dataset.}
\label{fig:result_freq_time_mtop}
\end{figure}

In this section, we present an analysis of the time-accuracy trade-off for our dynamic data pruning method in relation to $T$ and $\tau$ on the MTOP dataset, with a fixed prune rate of 70\%. Similar trends are observed with the other datasets, as shown in Appendix \ref{sec:further_results}. The results are illustrated in Figures \ref{fig:result_freq_mtop} and \ref{fig:result_freq_time_mtop}.

We found that training on larger $\tau$ and considering shorter $T$ results in higher accuracy. We note that $\tau$ is key to our approach since it impacts accuracy the most. For instance, a $\tau$ of 2 is less effective than 16, while the difference between 4 and 8 is marginal. Despite pruning 70\% of the data points, we achieved similar results to full training with $T=2$ and $\tau=16$. Considering that we are at a high prune rate, we can achieve a better time-accuracy trade-off at $T=4$ and $\tau=4$. This results in a marginal sacrifice of 1.2\% for twice the reduction in finetuning time. Specifically, the finetuning time is reduced from 826 s ($-36\%$) to 533 s ($-59\%$) compared to full training at 1295 s.

Furthermore, we observe the inverse relationship between the finetuning time and $T$ in Figure \ref{fig:result_freq_time_mtop} as described by equation \ref{eq:time_rho} in Appendix \ref{sec:time_proof}. For small $T$, this inverse relationship significantly increases the computational time. For instance, the difference between $T=1$ and $T=2$ is nearly 2 minutes, while being negligible between 8 and 16.

Due to periodical re-evaluation of the pruned subset, we note that performance obtained with $\tau$, $\rho$, and $T$ is less sensible to dataset characteristics, and the time-accuracy trade-off it provides remains robust across various settings. We provide additional recommendations about selecting these parameters in Appendix \ref{sec:param_selec}.

\section{Data Selection Analysis}
\label{sec:select_analysis}

We analyze pruning dynamics for joint intent classification and slot filling. The analysis is presented for SLURP, but observations are similar to other datasets.

\begin{figure}[ht!]
\includegraphics[width=\linewidth]{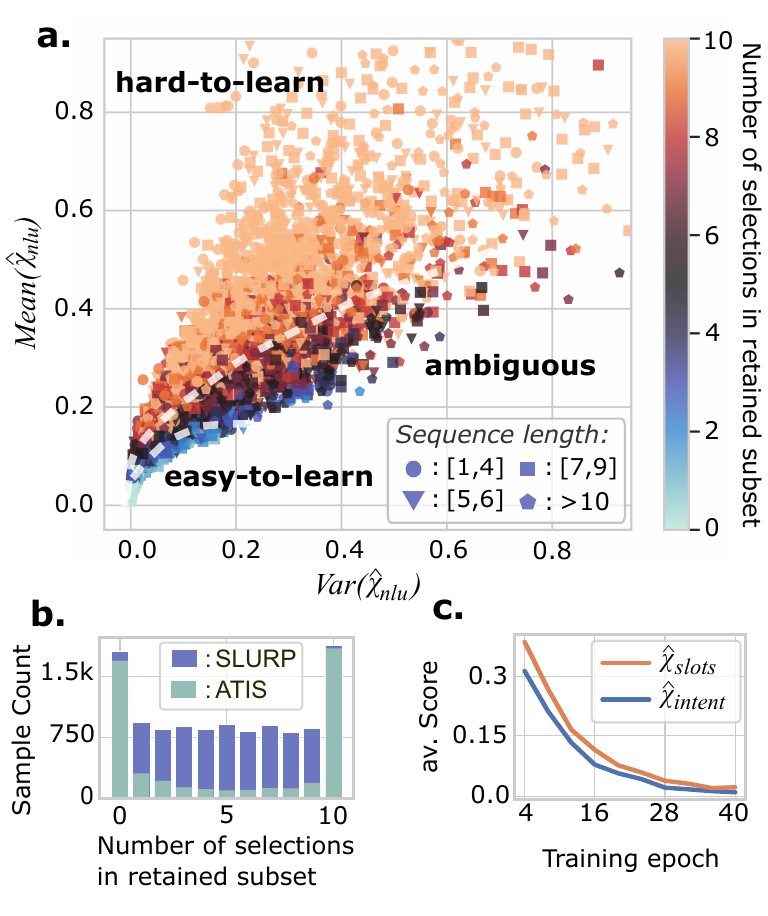}
\caption{\textit{a}. Data map for SLURP train set. Colours indicate the number of times an example is selected in the retained subset, and shapes show the sequence length. \textit{b}. Distribution of the number of selections in retained subset over ten pruning cycles for SLURP and ATIS. \textit{c}. Average intent and slots metrics with respect to epochs.}
\label{fig:datamap}
\end{figure}

First, we present a data map in Figure \ref{fig:datamap}\textit{a}, that shows variance and mean of $\hat{\chi}_{nlu}$ over 10 cycles of $T=4$ epochs for all training examples. The pruning proportion is set to 50\%. We discern three regions delimited by dashed curves: the bottom-left corner (low variance, low mean) corresponds to easy-to-learn examples consistently pruned from the train set, the top (high mean) corresponds to hard-to-learn examples consistently retained, and examples in-between (mid/high variance, low/mid mean) are ambiguous.

Figure \ref{fig:datamap}\textit{b} shows the distribution for the number of times a given example is selected in the retained subset over ten pruning cycles. For SLURP, we observe that about a third of the examples are consistently either always selected or never selected, corresponding to hard-to-learn and easy-to-learn cases, respectively. Other examples are approximately evenly distributed across selection counts. In comparison, we also present the distribution for the ATIS dataset, for which most training examples are either never selected or always selected.

Figure \ref{fig:datamap}\textit{c} presents scoring metrics averaged over the full train set for intent and slot classifications with respect to the training epoch. On average, sequences have one intent classification term per 6.9 slot classification terms. Overall, contribution to $\hat{\chi}_{nlu}$ is comparable between intent and slots, meaning that individual score is typically smaller for slots. However, scores associated with the prevalent \textit{no\_entity} class are relatively low compared to other slot labels, which have a predominant contribution to $\hat{\chi}_{slots}$.

In Table \ref{tab:datamap} of Appendix \ref{sec:further_results}, we present 13 examples of easy-to-learn, hard-to-learn and ambiguous samples, with their score averaged over time and the number of selections with four different initializations. Examples with the highest score typically have mislabeled intent (1) or slot (2), or have an infrequent class label (3, 0.03\% of all samples), or are unique (4). Examples with lowest $\hat{\chi}_{nlu}$ generally have short utterances with no entity label and with highly represented intent labels having many redundant tokens across class examples (see 5-7). Note that Spearman correlation between $\hat{\chi}_{nlu}$ and the number of words ($r=0.096$), number of entities ($r=0.205$) and intent class prevalence ($r=-0.109$) remains low when considering full train set.

Ambiguous examples, however, are more difficult to characterize qualitatively. We present several examples with an increasing scoring metric which are all under the same intent label for comparison purposes (see examples 8-13). There is no clear distinction between samples. In fact, when considering different initializations, the same ambiguous sample obtains different selection counts in the retained subset. This shows the necessity of using dynamic pruning: a static pruning algorithm will target easy-to-learn samples but will suffer from the variability of ambiguous samples. This high variability also explains the unexpectedly high performance of random pruning. Dynamic pruning can adjust the selection of ambiguous samples during training based on the model's performance, which leads to more efficient training.

When the proportion of ambiguous examples is low, as with the ATIS dataset, single and static pruning outperform random dynamic pruning, as shown in Figure \ref{fig:result_pruning_fullseq}, since the pruning subsets remain consistent across training.


\section{Conclusion}

In summary, this work presents a dynamic approach to data pruning to improve finetuning efficiency. It introduces an extension of the EL2N metric for multi-output classifications. We use this metric to periodically select a subset of the data during training in contrast to static pruning. We empirically show that our method provides a better trade-off between accuracy and efficiency, especially at high prune rates. We also show that this trade-off benefits from initial training with the full train set for a few epochs.

We apply our method to NLP classification datasets, particularly six tasks from the GLUE benchmark and four datasets for joint intent classification and slot filling. By pruning 50\% of training examples, we preserve full accuracy on the test set while reducing the finetuning time by up to 41\%. If we tolerate an accuracy reduction of 1\% absolute, we can prune 80\% of the training data, corresponding to time reductions of up to 66\%. Finally, we provide insights into how performance is affected by the characteristics of the training set.

As future work, we envision further investigations including more natural language tasks, more challenging datasets (e.g. SuperGLUE) and improved importance approximations. Overall, dynamical evaluation of sample importance remains largely unexplored beyond improving efficiency, notably in data augmentation, active learning, mislabel detection or pre-training data selection. 

\section*{Limitations}

This study is limited to classification tasks with an encoder architecture, short texts (e.g. utterances), datasets with at least several thousand examples, and a relatively low amount of mislabeled data. In theory, we could apply our method to longer texts, but our takeaways might not directly apply. For similar reasons, our conclusions could be different on very small datasets. Our approach is also sensible to mislabeled data points, but this weakness is mitigated by the fact that our method can contribute to improving the identification of such mislabels. We are also aware of further efficiency optimizations, such as calculating scores directly from mini-batches during training for the retained subset, which we leave to future work.

\section*{Ethics Statement}

We bring up two main ethical considerations. First, this empirical study uses a large language model requiring a considerable amount of computations (at most 1,500 GPU hours), which is not without environmental consequences. However, since this study aims at making training more efficient, it will help reduce energy consumption in the future. Moreover, this study focuses on accuracy as a measure of performance, which can hide pervasive effects on under-represented marginalized groups. However, since our method is about evaluating importance of training examples over train steps, it can lead to improving techniques to decrease bias in the training process, particularly when marginalized groups are under-represented in the data and therefore challenging to identify accurately.

\section*{Acknowledgements}
We would like to express our gratitude to François Beaulieu, Paul Vozila, Rupert Brook, and Alessandro Sordoni for their helpful comments. We also appreciate the valuable feedback provided by the anonymous reviewers, which contributed to the improvement of our paper.

\bibliography{anthology,custom}

\begin{thebibliography}{44}
\expandafter\ifx\csname natexlab\endcsname\relax\def\natexlab#1{#1}\fi

\bibitem[{Alain et~al.(2015)Alain, Lamb, Sankar, Courville, and
  Bengio}]{alain2015variance}
Guillaume Alain, Alex Lamb, Chinnadhurai Sankar, Aaron Courville, and Yoshua
  Bengio. 2015.
\newblock Variance reduction in sgd by distributed importance sampling.
\newblock \emph{arXiv preprint arXiv:1511.06481}.

\bibitem[{Ash et~al.(2019)Ash, Zhang, Krishnamurthy, Langford, and
  Agarwal}]{ash2019deep}
Jordan~T Ash, Chicheng Zhang, Akshay Krishnamurthy, John Langford, and Alekh
  Agarwal. 2019.
\newblock Deep batch active learning by diverse, uncertain gradient lower
  bounds.
\newblock In \emph{International Conference on Learning Representations}.

\bibitem[{Bastianelli et~al.(2020)Bastianelli, Vanzo, Swietojanski, and
  Rieser}]{bastianelli2020slurp}
Emanuele Bastianelli, Andrea Vanzo, Pawel Swietojanski, and Verena Rieser.
  2020.
\newblock Slurp: A spoken language understanding resource package.
\newblock In \emph{Proceedings of the 2020 Conference on Empirical Methods in
  Natural Language Processing (EMNLP)}, pages 7252--7262.

\bibitem[{Bengio et~al.(2009)Bengio, Louradour, Collobert, and
  Weston}]{bengio2009curriculum}
Yoshua Bengio, J{\'e}r{\^o}me Louradour, Ronan Collobert, and Jason Weston.
  2009.
\newblock Curriculum learning.
\newblock In \emph{Proceedings of the 26th annual international conference on
  machine learning}, pages 41--48.

\bibitem[{Campbell and Broderick(2018)}]{campbell2018bayesian}
Trevor Campbell and Tamara Broderick. 2018.
\newblock Bayesian coreset construction via greedy iterative geodesic ascent.
\newblock In \emph{International Conference on Machine Learning}, pages
  698--706. PMLR.

\bibitem[{Chen et~al.(2019)Chen, Zhuo, and Wang}]{chen2019bert}
Qian Chen, Zhu Zhuo, and Wen Wang. 2019.
\newblock Bert for joint intent classification and slot filling.
\newblock \emph{arXiv preprint arXiv:1902.10909}.

\bibitem[{Christopoulou et~al.(2022)Christopoulou, Lampouras, and
  Iacobacci}]{christopoulou2022training}
Fenia Christopoulou, Gerasimos Lampouras, and Ignacio Iacobacci. 2022.
\newblock \href {https://aclanthology.org/2022.emnlp-main.167} {Training
  dynamics for curriculum learning: A study on monolingual and cross-lingual
  {NLU}}.
\newblock In \emph{Proceedings of the 2022 Conference on Empirical Methods in
  Natural Language Processing}, pages 2595--2611, Abu Dhabi, United Arab
  Emirates. Association for Computational Linguistics.

\bibitem[{Coleman et~al.(2020)Coleman, Yeh, Mussmann, Mirzasoleiman, Bailis,
  Liang, Leskovec, and Zaharia}]{coleman2019selection}
C~Coleman, C~Yeh, S~Mussmann, B~Mirzasoleiman, P~Bailis, P~Liang, J~Leskovec,
  and M~Zaharia. 2020.
\newblock Selection via proxy: Efficient data selection for deep learning.
\newblock In \emph{International Conference on Learning Representations
  (ICLR)}.

\bibitem[{Coucke et~al.(2018)Coucke, Saade, Ball, Bluche, Caulier, Leroy,
  Doumouro, Gisselbrecht, Caltagirone, Lavril et~al.}]{coucke2018snips}
Alice Coucke, Alaa Saade, Adrien Ball, Th{\'e}odore Bluche, Alexandre Caulier,
  David Leroy, Cl{\'e}ment Doumouro, Thibault Gisselbrecht, Francesco
  Caltagirone, Thibaut Lavril, et~al. 2018.
\newblock Snips voice platform: an embedded spoken language understanding
  system for private-by-design voice interfaces.
\newblock \emph{arXiv preprint arXiv:1805.10190}.

\bibitem[{Devlin et~al.(2019)Devlin, Chang, Lee, and
  Toutanova}]{devlin2019bert}
Jacob Devlin, Ming-Wei Chang, Kenton Lee, and Kristina Toutanova. 2019.
\newblock Bert: Pre-training of deep bidirectional transformers for language
  understanding.
\newblock In \emph{Proceedings of NAACL-HLT}, pages 4171--4186.

\bibitem[{Fayyaz et~al.(2022)Fayyaz, Aghazadeh, Modarressi, Pilehvar,
  Yaghoobzadeh, Kahou, and Mila}]{fayyaz2022bert}
Mohsen Fayyaz, Ehsan Aghazadeh, Ali Modarressi, Mohammad~Taher Pilehvar,
  Yadollah Yaghoobzadeh, Samira~Ebrahimi Kahou, and CIFAR Mila. 2022.
\newblock Bert on a data diet: Finding important examples by gradient-based
  pruning.

\bibitem[{Gupta et~al.(2015)Gupta, Agrawal, Gopalakrishnan, and
  Narayanan}]{gupta2015deep}
Suyog Gupta, Ankur Agrawal, Kailash Gopalakrishnan, and Pritish Narayanan.
  2015.
\newblock Deep learning with limited numerical precision.
\newblock In \emph{International conference on machine learning}, pages
  1737--1746. PMLR.

\bibitem[{Har-Peled and Kushal(2005)}]{har2005smaller}
Sariel Har-Peled and Akash Kushal. 2005.
\newblock Smaller coresets for k-median and k-means clustering.
\newblock In \emph{Proceedings of the twenty-first annual symposium on
  Computational geometry}, pages 126--134.

\bibitem[{Katharopoulos and Fleuret(2018)}]{katharopoulos2018not}
Angelos Katharopoulos and Fran{\c{c}}ois Fleuret. 2018.
\newblock Not all samples are created equal: Deep learning with importance
  sampling.
\newblock In \emph{International conference on machine learning}, pages
  2525--2534. PMLR.

\bibitem[{Kim(2020)}]{kim2020deep}
Yekyung Kim. 2020.
\newblock Deep active learning for sequence labeling based on diversity and
  uncertainty in gradient.
\newblock In \emph{Proceedings of the 2nd Workshop on Life-long Learning for
  Spoken Language Systems}, pages 1--8.

\bibitem[{Kovaleva et~al.(2019)Kovaleva, Romanov, Rogers, and
  Rumshisky}]{kovaleva2019revealing}
Olga Kovaleva, Alexey Romanov, Anna Rogers, and Anna Rumshisky. 2019.
\newblock Revealing the dark secrets of bert.
\newblock In \emph{Proceedings of the 2019 Conference on Empirical Methods in
  Natural Language Processing and the 9th International Joint Conference on
  Natural Language Processing (EMNLP-IJCNLP)}.

\bibitem[{Kumar et~al.(2010)Kumar, Packer, and Koller}]{kumar2010self}
M~Kumar, Benjamin Packer, and Daphne Koller. 2010.
\newblock Self-paced learning for latent variable models.
\newblock \emph{Advances in neural information processing systems}, 23.

\bibitem[{Lhoest et~al.(2021)Lhoest, Villanova~del Moral, Jernite, Thakur, von
  Platen, Patil, Chaumond, Drame, Plu, Tunstall, Davison, {\v{S}}a{\v{s}}ko,
  Chhablani, Malik, Brandeis, Le~Scao, Sanh, Xu, Patry, McMillan-Major, Schmid,
  Gugger, Delangue, Matussi{\`e}re, Debut, Bekman, Cistac, Goehringer, Mustar,
  Lagunas, Rush, and Wolf}]{lhoest2021datasets}
Quentin Lhoest, Albert Villanova~del Moral, Yacine Jernite, Abhishek Thakur,
  Patrick von Platen, Suraj Patil, Julien Chaumond, Mariama Drame, Julien Plu,
  Lewis Tunstall, Joe Davison, Mario {\v{S}}a{\v{s}}ko, Gunjan Chhablani,
  Bhavitvya Malik, Simon Brandeis, Teven Le~Scao, Victor Sanh, Canwen Xu,
  Nicolas Patry, Angelina McMillan-Major, Philipp Schmid, Sylvain Gugger,
  Cl{\'e}ment Delangue, Th{\'e}o Matussi{\`e}re, Lysandre Debut, Stas Bekman,
  Pierric Cistac, Thibault Goehringer, Victor Mustar, Fran{\c{c}}ois Lagunas,
  Alexander Rush, and Thomas Wolf. 2021.
\newblock \href {https://doi.org/10.18653/v1/2021.emnlp-demo.21} {Datasets: A
  community library for natural language processing}.
\newblock In \emph{Proceedings of the 2021 Conference on Empirical Methods in
  Natural Language Processing: System Demonstrations}, pages 175--184, Online
  and Punta Cana, Dominican Republic. Association for Computational
  Linguistics.

\bibitem[{Li et~al.(2021)Li, Arora, Chen, Gupta, Gupta, and
  Mehdad}]{li2021mtop}
Haoran Li, Abhinav Arora, Shuohui Chen, Anchit Gupta, Sonal Gupta, and Yashar
  Mehdad. 2021.
\newblock Mtop: A comprehensive multilingual task-oriented semantic parsing
  benchmark.
\newblock In \emph{Proceedings of the 16th Conference of the European Chapter
  of the Association for Computational Linguistics: Main Volume}, pages
  2950--2962.

\bibitem[{Liu et~al.(2019)Liu, Ott, Goyal, Du, Joshi, Chen, Levy, Lewis,
  Zettlemoyer, and Stoyanov}]{liu2019roberta}
Yinhan Liu, Myle Ott, Naman Goyal, Jingfei Du, Mandar Joshi, Danqi Chen, Omer
  Levy, Mike Lewis, Luke Zettlemoyer, and Veselin Stoyanov. 2019.
\newblock Roberta: A robustly optimized bert pretraining approach.
\newblock \emph{arXiv preprint arXiv:1907.11692}.

\bibitem[{Micikevicius et~al.(2018)Micikevicius, Narang, Alben, Diamos, Elsen,
  Garcia, Ginsburg, Houston, Kuchaiev, Venkatesh
  et~al.}]{micikevicius2017mixed}
Paulius Micikevicius, Sharan Narang, Jonah Alben, Gregory Diamos, Erich Elsen,
  David Garcia, Boris Ginsburg, Michael Houston, Oleksii Kuchaiev, Ganesh
  Venkatesh, et~al. 2018.
\newblock Mixed precision training.
\newblock In \emph{International Conference on Learning Representations}.

\bibitem[{Mindermann et~al.(2022)Mindermann, Brauner, Razzak, Sharma, Kirsch,
  Xu, H{\"o}ltgen, Gomez, Morisot, Farquhar et~al.}]{mindermann2022prioritized}
S{\"o}ren Mindermann, Jan~M Brauner, Muhammed~T Razzak, Mrinank Sharma, Andreas
  Kirsch, Winnie Xu, Benedikt H{\"o}ltgen, Aidan~N Gomez, Adrien Morisot,
  Sebastian Farquhar, et~al. 2022.
\newblock Prioritized training on points that are learnable, worth learning,
  and not yet learnt.
\newblock In \emph{International Conference on Machine Learning}, pages
  15630--15649. PMLR.

\bibitem[{Mishra and Sachdeva(2020)}]{mishra2020we}
Swaroop Mishra and Bhavdeep~Singh Sachdeva. 2020.
\newblock Do we need to create big datasets to learn a task?
\newblock In \emph{Proceedings of SustaiNLP: Workshop on Simple and Efficient
  Natural Language Processing}, pages 169--173.

\bibitem[{Munteanu et~al.(2018)Munteanu, Schwiegelshohn, Sohler, and
  Woodruff}]{munteanu2018coresets}
Alexander Munteanu, Chris Schwiegelshohn, Christian Sohler, and David Woodruff.
  2018.
\newblock On coresets for logistic regression.
\newblock \emph{Advances in Neural Information Processing Systems}, 31.

\bibitem[{Park et~al.()Park, Papailiopoulos, and Lee}]{park2022active}
Dongmin Park, Dimitris Papailiopoulos, and Kangwook Lee.
\newblock Active learning is a strong baseline for data subset selection.
\newblock In \emph{Has it Trained Yet? NeurIPS 2022 Workshop}.

\bibitem[{Paszke et~al.(2017)Paszke, Gross, Chintala, Chanan, Yang, DeVito,
  Lin, Desmaison, Antiga, and Lerer}]{paszke2017automatic}
Adam Paszke, Sam Gross, Soumith Chintala, Gregory Chanan, Edward Yang, Zachary
  DeVito, Zeming Lin, Alban Desmaison, Luca Antiga, and Adam Lerer. 2017.
\newblock Automatic differentiation in pytorch.

\bibitem[{Paszke et~al.(2019)Paszke, Gross, Massa, Lerer, Bradbury, Chanan,
  Killeen, Lin, Gimelshein, Antiga, Desmaison, Kopf, Yang, DeVito, Raison,
  Tejani, Chilamkurthy, Steiner, Fang, Bai, and Chintala}]{paszke2019pytorch}
Adam Paszke, Sam Gross, Francisco Massa, Adam Lerer, James Bradbury, Gregory
  Chanan, Trevor Killeen, Zeming Lin, Natalia Gimelshein, Luca Antiga, Alban
  Desmaison, Andreas Kopf, Edward Yang, Zachary DeVito, Martin Raison, Alykhan
  Tejani, Sasank Chilamkurthy, Benoit Steiner, Lu~Fang, Junjie Bai, and Soumith
  Chintala. 2019.
\newblock \href
  {http://papers.neurips.cc/paper/9015-pytorch-an-imperative-style-high-performance-deep-learning-library.pdf}
  {Pytorch: An imperative style, high-performance deep learning library}.
\newblock In \emph{Advances in Neural Information Processing Systems 32}, pages
  8024--8035. Curran Associates, Inc.

\bibitem[{Paul et~al.(2021)Paul, Ganguli, and Dziugaite}]{paul2021deep}
Mansheej Paul, Surya Ganguli, and Gintare~Karolina Dziugaite. 2021.
\newblock Deep learning on a data diet: Finding important examples early in
  training.
\newblock \emph{Advances in Neural Information Processing Systems},
  34:20596--20607.

\bibitem[{Platanios et~al.(2019)Platanios, Stretcu, Neubig, Pocz{\'o}s, and
  Mitchell}]{platanios2019competence}
Emmanouil~Antonios Platanios, Otilia Stretcu, Graham Neubig, Barnab{\'a}s
  Pocz{\'o}s, and Tom Mitchell. 2019.
\newblock Competence-based curriculum learning for neural machine translation.
\newblock In \emph{Proceedings of the 2019 Conference of the North American
  Chapter of the Association for Computational Linguistics: Human Language
  Technologies, Volume 1 (Long and Short Papers)}, pages 1162--1172.

\bibitem[{Price(1990)}]{price1990atis}
Patti Price. 1990.
\newblock Evaluation of spoken language systems: The atis domain.
\newblock In \emph{Speech and Natural Language: Proceedings of a Workshop Held
  at Hidden Valley, Pennsylvania, June 24-27, 1990}.

\bibitem[{Raju et~al.(2021)Raju, Daruwalla, and Lipasti}]{raju2021accelerating}
Ravi~S Raju, Kyle Daruwalla, and Mikko Lipasti. 2021.
\newblock Accelerating deep learning with dynamic data pruning.
\newblock \emph{arXiv preprint arXiv:2111.12621}.

\bibitem[{Sorscher et~al.(2022)Sorscher, Geirhos, Shekhar, Ganguli, and
  Morcos}]{sorscher2022beyond}
Ben Sorscher, Robert Geirhos, Shashank Shekhar, Surya Ganguli, and Ari~S
  Morcos. 2022.
\newblock Beyond neural scaling laws: beating power law scaling via data
  pruning.
\newblock In \emph{Advances in Neural Information Processing Systems}.

\bibitem[{Swayamdipta et~al.(2020)Swayamdipta, Schwartz, Lourie, Wang,
  Hajishirzi, Smith, and Choi}]{swayamdipta2020dataset}
Swabha Swayamdipta, Roy Schwartz, Nicholas Lourie, Yizhong Wang, Hannaneh
  Hajishirzi, Noah~A Smith, and Yejin Choi. 2020.
\newblock Dataset cartography: Mapping and diagnosing datasets with training
  dynamics.
\newblock In \emph{Proceedings of the 2020 Conference on Empirical Methods in
  Natural Language Processing (EMNLP)}, pages 9275--9293.

\bibitem[{Tamkin et~al.(2020)Tamkin, Singh, Giovanardi, and
  Goodman}]{tamkin2020investigating}
Alex Tamkin, Trisha Singh, Davide Giovanardi, and Noah Goodman. 2020.
\newblock Investigating transferability in pretrained language models.
\newblock In \emph{Findings of the Association for Computational Linguistics:
  EMNLP 2020}, pages 1393--1401.

\bibitem[{Toneva et~al.(2018)Toneva, Sordoni, des Combes, Trischler, Bengio,
  and Gordon}]{toneva2018empirical}
Mariya Toneva, Alessandro Sordoni, Remi~Tachet des Combes, Adam Trischler,
  Yoshua Bengio, and Geoffrey~J Gordon. 2018.
\newblock An empirical study of example forgetting during deep neural network
  learning.
\newblock In \emph{International Conference on Learning Representations}.

\bibitem[{Tur et~al.(2010)Tur, Hakkani-T{\"u}r, and Heck}]{tur2010left}
Gokhan Tur, Dilek Hakkani-T{\"u}r, and Larry Heck. 2010.
\newblock What is left to be understood in atis?
\newblock In \emph{2010 IEEE Spoken Language Technology Workshop}, pages
  19--24. IEEE.

\bibitem[{Wan et~al.(2020)Wan, Yang, Wong, Zhou, Chao, Zhang, and
  Chen}]{wan2020self}
Yu~Wan, Baosong Yang, Derek~F Wong, Yikai Zhou, Lidia~S Chao, Haibo Zhang, and
  Boxing Chen. 2020.
\newblock Self-paced learning for neural machine translation.
\newblock In \emph{Proceedings of the 2020 Conference on Empirical Methods in
  Natural Language Processing (EMNLP)}, pages 1074--1080.

\bibitem[{Wang et~al.(2019)Wang, Singh, Michael, Hill, Levy, and
  Bowman}]{wang2019glue}
Alex Wang, Amanpreet Singh, Julian Michael, Felix Hill, Omer Levy, and Samuel~R
  Bowman. 2019.
\newblock Glue: A multi-task benchmark and analysis platform for natural
  language understanding.
\newblock In \emph{7th International Conference on Learning Representations,
  ICLR 2019}.

\bibitem[{Wolf et~al.(2020)Wolf, Debut, Sanh, Chaumond, Delangue, Moi, Cistac,
  Rault, Louf, Funtowicz et~al.}]{wolf2020transformers}
Thomas Wolf, Lysandre Debut, Victor Sanh, Julien Chaumond, Clement Delangue,
  Anthony Moi, Pierric Cistac, Tim Rault, R{\'e}mi Louf, Morgan Funtowicz,
  et~al. 2020.
\newblock Transformers: State-of-the-art natural language processing.
\newblock In \emph{Proceedings of the 2020 conference on empirical methods in
  natural language processing: system demonstrations}, pages 38--45.

\bibitem[{Xiong et~al.(2021)Xiong, Zeng, Chakraborty, Tan, Fung, Li, and
  Singh}]{xiong2021nystromformer}
Yunyang Xiong, Zhanpeng Zeng, Rudrasis Chakraborty, Mingxing Tan, Glenn Fung,
  Yin Li, and Vikas Singh. 2021.
\newblock Nystr{\"o}mformer: A nystr{\"o}m-based algorithm for approximating
  self-attention.
\newblock In \emph{Proceedings of the AAAI Conference on Artificial
  Intelligence}, volume~35, pages 14138--14148.

\bibitem[{Yang et~al.(2022)Yang, Xie, Peng, Xu, Sun, and
  Li}]{yang2022influence}
Shuo Yang, Zeke Xie, Hanyu Peng, Min Xu, Mingming Sun, and Ping Li. 2022.
\newblock Dataset pruning: Reducing training data by examining generalization
  influence.
\newblock \emph{arXiv preprint arXiv:2205.09329}.

\bibitem[{Zaheer et~al.(2020)Zaheer, Guruganesh, Dubey, Ainslie, Alberti,
  Ontanon, Pham, Ravula, Wang, Yang et~al.}]{zaheer2020big}
Manzil Zaheer, Guru Guruganesh, Kumar~Avinava Dubey, Joshua Ainslie, Chris
  Alberti, Santiago Ontanon, Philip Pham, Anirudh Ravula, Qifan Wang, Li~Yang,
  et~al. 2020.
\newblock Big bird: Transformers for longer sequences.
\newblock \emph{Advances in Neural Information Processing Systems},
  33:17283--17297.

\bibitem[{Zayed et~al.(2022)Zayed, Parthasarathi, Mordido, Palangi, Shabanian,
  and Chandar}]{zayed2022deep}
Abdelrahman Zayed, Prasanna Parthasarathi, Goncalo Mordido, Hamid Palangi,
  Samira Shabanian, and Sarath Chandar. 2022.
\newblock Deep learning on a healthy data diet: Finding important examples for
  fairness.
\newblock \emph{arXiv preprint arXiv:2211.11109}.

\bibitem[{Zhang and He(2020)}]{zhang2020accelerating}
Minjia Zhang and Yuxiong He. 2020.
\newblock Accelerating training of transformer-based language models with
  progressive layer dropping.
\newblock \emph{Advances in Neural Information Processing Systems},
  33:14011--14023.

\end{thebibliography}
\bibliographystyle{acl_natbib}

\appendix

\section{Derivations of Pruning Metrics}
\label{sec:el2n_equations}
\subsection{Unique Single-Label Classification on CLS token}

We begin by defining the GraNd score, $\chi_t(x,y) \in \mathbb{R}_{\geq0}$, at time $t > 0$ and for any sample data $(x,y)$ as a measure of sample importance \citep{paul2021deep}:

\begin{equation}
\label{eq:grand}
\chi_t(x,y) = \mathrm{E}_{\mathbf{w}}\left[||g_t(x,y)||_2\right],
\end{equation}

where $g_t$ is the gradient of the sample $(x,y)$ obtained with model $f$ at time $t$. For simplicity, we define $\chi'_t(x,y)$ as the same expression as in Equation \eqref{eq:grand} but without the expectation term. We compensate for the improvement of this expectation with dynamic pruning by leveraging the exponential moving average:

\begin{equation}
\chi'_t(x,y) = ||g_t(x,y)||_2.
\end{equation}

Because processing per-sample gradients can be computationally burdensome, we look for an upper bound. First, we use the result from \citet{katharopoulos2018not}, which implies that the full gradient is upper bounded by a constant $\beta \in \mathbb R_{>0}$ multiplied by the gradient of the last classification layer. Let $G_t(x,y) \in \mathbb{R}^{K \times d}$ denote the gradient of the last classification layer, where $K,d \in \mathbb{N}$ (number of classes and hidden dimensions, respectively). We can write this result as

\begin{equation}
\label{eq:bound_grand}
\chi'_t(x,y) \leq \beta ||G_t(x,y)||_2.
\end{equation}

Since the last layer is a classification with the cross-entropy loss on a softmax activation, it naturally takes the form of a $K \times d$ matrix defined by the outer product of the error vector $\vec{p}(x) - \vec{y}$ and the transposed contextual word embedding $\vec{h}_{CLS}(x) \in \mathbb{R}^{d}$. For simplification purposes, we do not carry the notation marking the dependence on $x$ for $\vec{p}$ and $\vec{h}_{CLS}$:

\begin{equation}
\label{eq:last_layer_grad}
G_t(x,y) = \left(\vec{p} - \vec{y}\right) \vec{h}_{CLS}^T.
\end{equation}

By replacing Equation \eqref{eq:last_layer_grad} in Equation \eqref{eq:bound_grand}, we obtain

\begin{equation}
\chi'_t(x,y) \leq \beta ||\left(\vec{p} - \vec{y}\right) \vec{h}_{CLS}^T||_2.
\end{equation}

Using the property of outer products, we can separate the matrix norm into the product of the vector norms:

\begin{equation}
\chi'_t(x,y) \leq \beta ||\vec{h}_{CLS}||_2 ||\vec{p} - \vec{y}||_2.
\end{equation}

Given that we want to rank samples $\{(x,y)\}_{1..N}$ and that norm of the contextual word embeddings of $\vec{h}_{CLS}$ can roughly be assumed constant across samples due to layer normalization, we obtain:

\begin{equation}
    \chi'_t(x,y) \propto ||\vec{p} - \vec{y}||_2
\end{equation}

This metric corresponds to the EL2N score proposed by \citet{paul2021deep}, and provides scoring function for intent classification:

\begin{equation}
    \hat{\chi}_{intent}(x,y) = ||\vec{p}(x) - \vec{y}||_2
\end{equation}

\subsection{Sequence of Single-Label Classifications}
\label{sec:sequence_el2n}

The derivation for multi-output classification is similar to that of single-label classification shown above, with some modifications. In this case, we consider a summation over the sequence length $m \in [1,M]$, where $M \in \mathbb{N}$, to update the entity classifier with the sum of the gradients from all tokens. We assume that all sequences are padded up to length $M$.

To begin, we introduce the sequence-wise error vectors $\vec{\delta}_{m}=\vec{p}_m - \vec{y_m}$, and rewrite the gradient $G_t(x,y)$ of equation \eqref{eq:last_layer_grad} as follows:

\begin{equation}
\label{eq:last_layer_grad_seq}
G_t(x,y) = \sum_{m=1}^M \vec{\delta}_{m} \vec{h}_{m}^T,
\end{equation}

where $\vec{h}_m$ is the contextual embedding for the $m$-th token. By substitution in equation \eqref{eq:bound_grand}, we obtain the following upper bound:

\begin{equation}
\chi'_t(x,y) \leq \beta ||\sum_{m=1}^M \vec{\delta}_{m} \vec{h}_{m}^T||_2.
\end{equation}

We then use the triangle inequality to derive a new bound as follows:

\begin{equation}
\chi'_t(x,y) \leq \beta \sum_{m=1}^M ||\vec{\delta}_{m} \vec{h}_{m}^T||_2.
\end{equation}

Since the sum over $m \in [1,M]$ can be interpreted as a $\ell_1$ norm over the sequence, we replace it with a $\ell_2$ norm leveraging the equivalence of topology of $p$-norms. This norm is less sensitive to the sequence length and puts more emphasis on the highest norms. This results in a new upper bound:

\begin{equation}
\chi'_t(x,y) \leq \beta \sqrt{ M \sum_{m=1}^M ||\vec{\delta}_{m} \vec{h}_{m}^T||_2^2}.
\end{equation}

This substitution is better suited for our metric for two reasons. First, the square applied to each individual norm of element $m$ reduces the impact of small norms compared to larger ones, resulting in a metric that scales less with the length of the sequence $M$. Second, it matches the $\ell_2$ norm used in the previous section, which combines more naturally in the mathematical development of the next section for joint tasks.

We can apply the inner $\ell_2$ norm separately on the outer product of vectors. Since the amplitude of all contextual word embeddings $\vec{h}_{m}$ are similar, we can assume $||\vec{h}_{m}||\approx||\vec{h}_{m'}||\approx||\vec{h}||$ with $m' \in \left[1,M\right]$ across tokens. Thus, we obtain the following simplified bound:

\begin{equation}
\chi'_t(x,y) \leq \beta \sqrt{M} ||\vec{h}||_2 \sqrt{\sum_{m=1}^M||\vec{\delta}_{m}||_2^2}.
\end{equation}

Since we are interested in ranking samples $\{(x,y)\}_{1..N}$ by scores as before, we can simplify this bound for multi-output classification tasks such as slot filling. We also replace $\vec{\delta}_{m}$ by its definition.

\begin{equation}
    \hat{\chi}_{slot}(x,y) = \sqrt{\sum_{m=1}^M||\vec{p}_m(x) - \vec{y}_m||_2^2}
\end{equation}

\subsection{Joint Single-Label Classification and Sequence of Single-Label Classification}

We aim to combine the gradient matrices of two independent classification heads into one gradient matrix using an $\ell_2$ norm. We can use the direct sum to express this combination for a joint task, such as joint intent classification (indicated by the \textit{int} superscript) and slot filling (indicated by the \textit{slot} superscript). The direct sum of matrices generates a block diagonal matrix that combines two vector spaces. Under a norm, this matrix representation is equivalent to the flattened one.

Let us denote the gradient matrices as $G^{(int,t)}(x,y_{int})$ and $G^{(slot,t)}(x,y_{slot})$, where $t$ denotes the training step, $x$ is the input data, and $y_{int}$ and $y_{slot}$ is the corresponding labels. For simplification, we are leaving the dependence on $x$, $y_{int}$ and $y_{slot}$ on the right side of our equations. The combined gradient matrix is obtained by taking the direct sum of these matrices as follows:

\begin{equation}
G_t(x,y) = G^{(int,t)} \oplus G^{(slot,t)}.
\end{equation}

Substituting this equation in \eqref{eq:bound_grand}, we get:

\begin{equation}
\chi'_t(x,y) \leq \beta || G^{(int,t)} \oplus G^{(slot,t)} ||_2.
\end{equation}

We consider the squared Frobenius norm and apply the sum of square terms block-wise, where the off-diagonal zero terms are ignored. This gives us the sum of the square norms of each gradient matrix as follows:

\begin{equation}
\small{
    \chi'_t(x,y)^2 \leq \beta^2 \left( ||G^{(int,t)}||_2^2 + ||G^{(slot,t)}||_2^2 \right).
}
\end{equation}

Using the definitions of the gradient norms obtained from the previous derivations, we get the following:

\begin{equation}
\begin{split}
\chi'_t(x,y)^2 &\leq \beta^2 ||\vec{h}_{CLS}||_2^2||\vec{p}_{int} - \vec{y}_{int}||_2^2 + \\
&\beta^2 ||\vec{h}||_2^2\sum_{m=1}^M||\vec{p}_m - \vec{y}_m||_2^2.
\end{split}
\end{equation}

As $||\vec{h}_{CLS}||\approx||\vec{h}||$ and by identification with the two previous derivations, we obtain the final form for the joint task, such as joint intent classification and slot filling, as follows:

\begin{equation}
   \hat{\chi}_{nlu}(x,y) = \sqrt{\hat{\chi}_{intent}^2  + \hat{\chi}_{slot}^2 }.
\end{equation}

\section{Derivations of GPU Runtime Equations}
\label{sec:time_proof}

\subsection{GPU Runtime Dependency to Pruning Factor and Epoch Cycle}

The total GPU time $\mathcal{T}$ contains three terms as follow

\begin{equation}
\label{eq:sum_t}
    \mathcal{T} = \mathcal{T}_{init} + \mathcal{T}_{prune} + \mathcal{T}_{score},
\end{equation}

where $\mathcal{T}_{init}$ is the initial amount of epochs finetuned with the full training set, $\mathcal{T}_{prune}$ is the remaining amount of epochs to finetune with the pruned dataset, and $\mathcal{T}_{score}$ is the time to compute the importance score.

We define $\tau \leq E \in \mathbb{N}$, the epoch at which we start the data pruning and the total amount of epochs, respectively. We have $B \in \mathbb{N}$, the number of steps in one epoch (i.e. the ceiling of training set length divided by the batch size). We define an epoch cycle to prune the train data set as well $T \in \mathbb{N}$, with $T \leq E - \tau$. The proportion of pruned samples is defined by $\rho \in \left(0, 1\right)$. Finally, we have $\Delta t_{step} \in \mathbb{R}_{\geq0}$, the average time to process one mini-batch, and $\Delta t_{forward} \in \mathbb{R}_{\geq0}$, the time to compute a complete forward pass on the training dataset.

Then, we can explicitly formulate

\begin{equation}
    \mathcal{T}_{init} = \tau B \Delta t_{step}
\end{equation}

\begin{equation}
    \mathcal{T}_{prune} = \left(E-\tau\right) \left(1-\rho\right) B \Delta t_{step}
\end{equation}

\begin{equation}
    \mathcal{T}_{score} = \left\lfloor\frac{E-\tau}{T}\right\rfloor \Delta t_{forward},
\end{equation}

where $\left\lfloor\cdot\right\rfloor$ is the floor operator. Therefore, the complete GPU runtime equation based on eq. \ref{eq:sum_t} takes the form

\begin{equation}
\label{eq:time_rho}
    \begin{split}
    \mathcal{T} = E B \Delta t_{step} - \left(E-\tau\right) B \Delta t_{step} \rho \\
        + \left\lfloor\frac{E-\tau}{T}\right\rfloor \Delta t_{forward}
    \end{split}
\end{equation}

\subsection{Lower-Bound Epoch Cycle}

We can find $T_{min}$, which is a lower bound to $T$, to achieve an effective pruning in terms of computational time. We need to satisfy:

\begin{equation}
    \mathcal{T} < \mathcal{T}_{baseline}
\end{equation}

Given the previous definitions, we can find 

\begin{equation}
\label{eq:time_baseline}
    \mathcal{T}_{baseline} = E B \Delta t_{step}
\end{equation}

Considering eq. \ref{eq:time_baseline} and \ref{eq:time_rho} for $\mathcal{T}$, we deduce

\begin{equation}
\label{eq:cycle_bound}
    T_{min} > \frac{\Delta t_{forward}}{\Delta t_{step} B \rho}
\end{equation}

\section{Training Setup}
\label{sec:train_setup}

We use the following HuggingFace libraries distributed under the Apache License 2.0: transformers v4.21.3 \cite{wolf2020transformers} and datasets v2.5.1 \cite{lhoest2021datasets}. We set our back-end to Pytorch v1.10.1 \cite{paszke2017automatic,paszke2019pytorch}. We used the Adam optimizer ($\beta_1=0.9$, $\beta_2=0.999$ and $\epsilon=1\times10^{-8}$) with no warm up and no scheduler. We run fine-tunings and evaluations on an on-premise cluster of NVIDIA V100 32 Gb GPUs. All the results are the average of 5 runs with different random seeds unless it is stated otherwise.

We run our fine-tuning on the GLUE datasets with the following hyper-parameters: number of epochs of $10$, a learning rate of $2\times10^{-5}$, a batch size of $32$ and a maximum sequence length of $128$ (except for the RTE dataset at $256$). For the joint intent and slot filling datasets, we set: number of epochs of $40$, a learning rate of $2\times10^{-5}$, a batch size of $32$, a maximum sequence length of $50$ and $\lambda=0.5$. For $\hat{\chi}_{ema}$ calculations, we use $\alpha=0.8$.

\section{Dataset Statistics}
\label{sec:data_stat}

In Tables \ref{tab:data_glue_stats} and \ref{tab:data_nlu_stats}, we provide the statistics about the GLUE datasets and the joint NLU datasets. We verify that all samples in the datasets do not contain offensive language and personal information.


\begingroup
\setlength{\tabcolsep}{4pt}
\begin{table}[h]
\small
\caption{Statistics of GLUE datasets.}
\begin{tabular}{lrrrrrr}
\toprule
\textbf{Name} & |\textbf{Train}| & |\textbf{Test}| & \textbf{Task} & \textbf{Domain} \\
\midrule
COLA  & 8.5k & 1k & Acceptability & Misc. \\
MNLI & 393k & 20k & NLI & Misc. \\
MRPC & 3.7k & 1.7k & Paraphrase & News \\
QQP & 364k & 391k & Paraphrase & QA questions \\
RTE & 2.5k & 3k & NLI & News, Wikipedia \\
SST2 & 67k & 1.8k & Sentiment & Movie reviews\\
\bottomrule
\end{tabular}
\label{tab:data_glue_stats}
\end{table}
\endgroup

\begingroup
\setlength{\tabcolsep}{4pt}
\begin{table}[h]
\small
\caption{Statistics of joint NLU Datasets.}
\begin{tabular}{lrrrrr}
\toprule
\textbf{Name} & |\textbf{Train}| & |\textbf{Test}| & \textbf{Domains} & \textbf{Intents} & \textbf{Slots} \\
\midrule
ATIS  & 5.0k & 893 & 1 & 26 & 129 \\
SNIPS & 13.0k & 700 & - & 7 & 53 \\
SLURP & 16.5k & 3.0k & 18  & 60 & 55 \\
MTOP & 15.7k & 4.4k & 11 & 117 & 78 \\
\bottomrule
\end{tabular}
\label{tab:data_nlu_stats}
\end{table}
\endgroup


\section{Parameter Selection}
\label{sec:param_selec}

Values for $\tau$, $\rho$, and $T$ must be selected to achieve the desired time-accuracy trade-off. A valid first step is to set $\tau$ at about 10\% of $E$ for datasets with more than a few thousand samples or 20-40\% for smaller datasets to achieve higher accuracy. Larger $\tau$ significantly hinders efficiency at the expense of accuracy. Then, we calculate $T_{min}$ as a lower bound to obtain an efficiency gain with a given prune rate $\rho$ using equation \ref{eq:cycle_bound}. We recommend $\rho \in [0.4, 0.7]$ depending if accuracy or efficiency is more important, respectively. Finally, we want to set $T$ such that it is above $T_{min}$ to be efficient but small enough to provide many pruning updates for higher accuracy --- about ten cycles calculated with $\left\lfloor\frac{E-\tau}{T}\right\rfloor$ being a valid target. In Table \ref{tab:glue_w_results} of Appendix \ref{sec:further_results}, we present values of $T_{min}$ for $\rho \in \{0.1,0.5\}$ for all our datasets, along with the forward pass time $\Delta t_{f}$, train step time $\Delta t_{step}$ and the number of steps per epoch $B$.

\section{Further Results}
\label{sec:further_results}

In Table \ref{tab:glue_w_results}, we show $T_{min}$ (i.e. the minimum cycle to achieve an efficiency improvement over the baseline) for $\rho \in \left\{0.1,0.5\right\}$ for all datasets along their $\Delta t_{forward}$ and $\Delta t_{step}$.

\begingroup
\setlength{\tabcolsep}{4.5pt}
\begin{table}[ht]
    \centering
    \caption{Minimum epoch per prune cycle $T\left(\rho\right)=T_{min}\left(\rho\right)$ for prune rates of 10\% ($0.1$) and 50\% ($0.5$). $\Delta t_{f}$ is equivalent to $\Delta t_{forward}$. We measure $\Delta t_{f}$ and $\Delta t_{step}$ based on the median of previous experiments. $B$ is calculated from the train set length and the batch size.}
    \small
    \begin{tabular}{llrrrrr}
    \toprule
    {} & {} & $\Delta t_{f}$ & $\Delta t_{step}$ & $B$ & $T\left(0.1\right)$ & $T\left(0.5\right)$ \\
    \midrule
    \parbox[t]{1mm}{\multirow{6}{*}{\rotatebox[origin=c]{90}{GLUE}}} & COLA & 1.8 & 0.061 & 268 & 1.1 & 0.2 \\
    & MNLI & 145.4 & 0.082 & 12272 & 1.4 &  0.3 \\
    & MRPC & 1.8 & 0.076 & 115 & 2.0 & 0.4 \\
    & QQP & 112.4 & 0.068 & 11371 & 1.4 & 0.3 \\
    & RTE & 1.5 & 0.102 & 78 & 1.8 & 0.4 \\
    & SST2 & 14.5 & 0.062 & 2105 & 1.1 & 0.2 \\
    \midrule
    \parbox[t]{1mm}{\multirow{4}{*}{\rotatebox[origin=c]{90}{Joint NLU}}} & ATIS & 3.7 & 0.065 & 156 & 3.6 & 0.7 \\
    & MTOP & 8.3 & 0.065 & 490 & 2.6 & 0.5 \\
    & SLURP & 5.9 & 0.066 & 360 & 2.5 & 0.5 \\
    & SNIPS & 7.6 &   0.064 & 409 & 2.9 & 0.6 \\
    \bottomrule
    \end{tabular}
    \label{tab:glue_w_results}
\end{table}
\endgroup

\begin{figure*}[ht!]
\includegraphics[width=0.95\linewidth]{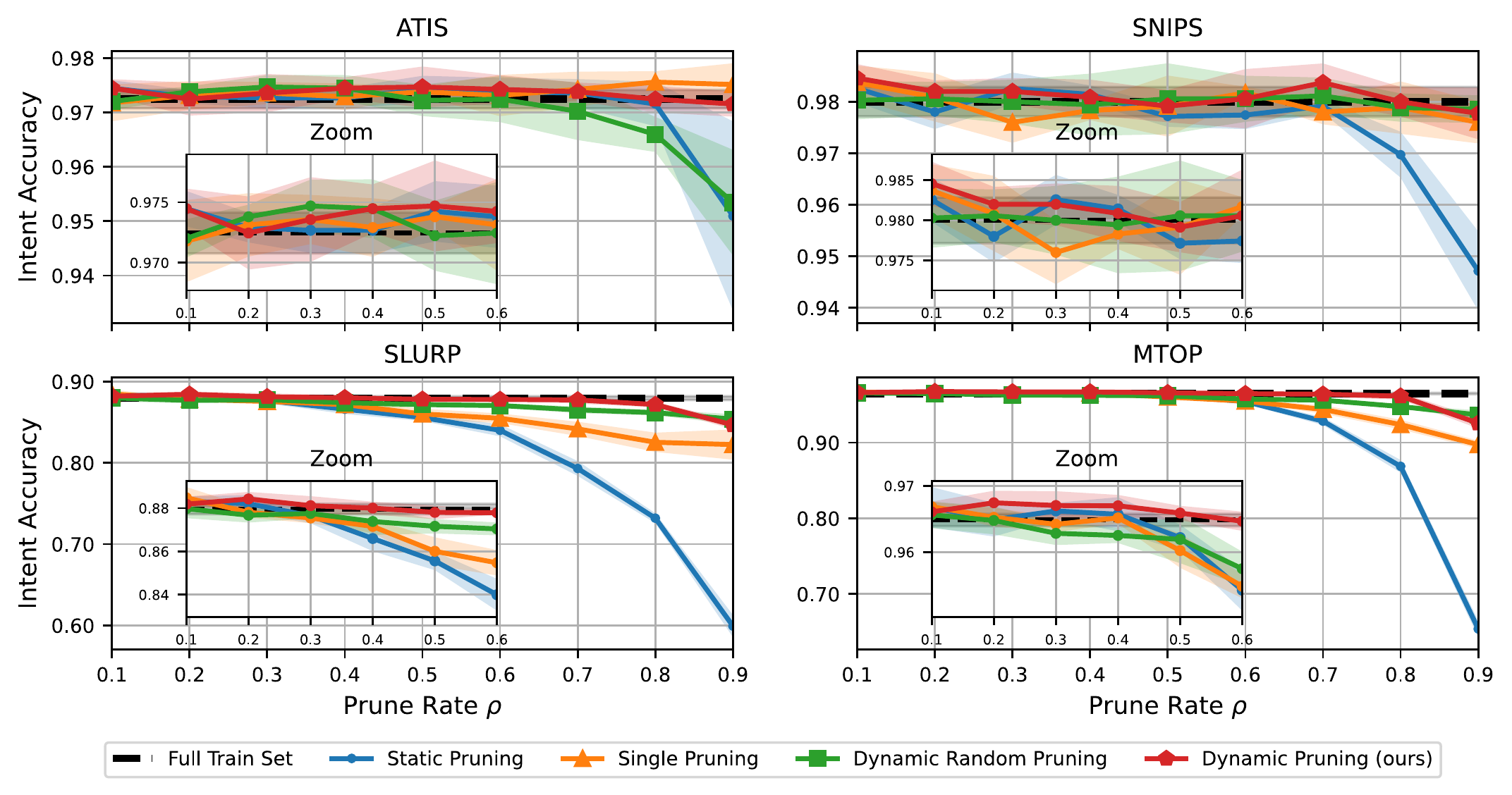}
\caption{Intent accuracy achieved on 40 epochs for different prune rates applying: static pruning (EL2N from 10 runs of 10 epochs), single pruning, dynamic random pruning, and our dynamic pruning (EL2N with EMA). The dynamic methods are run with $\tau=4$ and $T=4$.}
\label{fig:result_pruning_intent}
\end{figure*}

\begin{figure*}[ht!]
\includegraphics[width=0.95\linewidth]{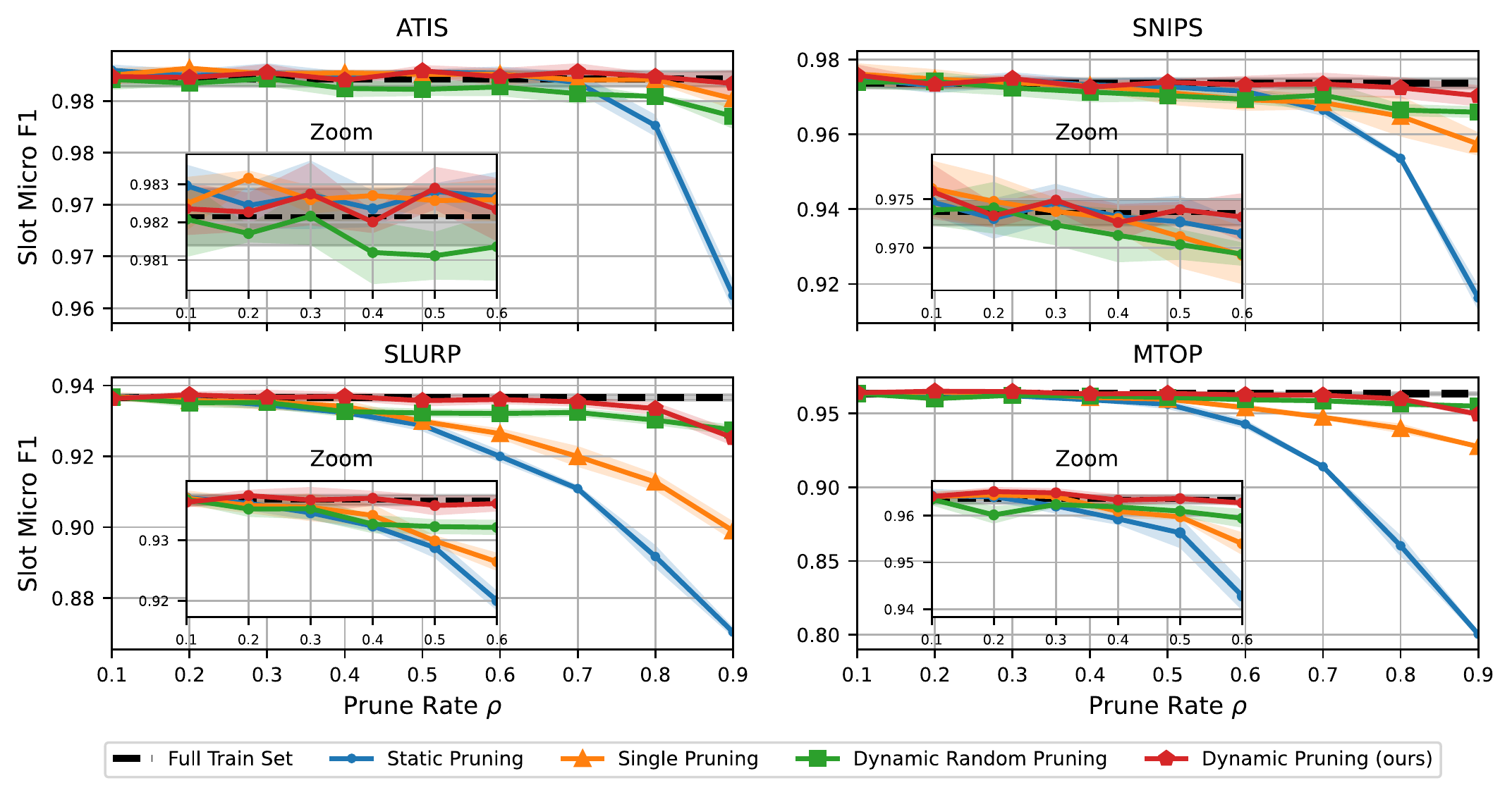}
\caption{Slot Micro F1 score achieved on 40 epochs for different prune rates applying: static pruning (EL2N from 10 runs of 10 epochs), single pruning, dynamic random pruning, and our dynamic pruning (EL2N with EMA). The dynamic methods are run with $\tau=4$ and $T=4$.}
\label{fig:result_pruning_slot}
\end{figure*}

In Figures \ref{fig:result_pruning_intent} and \ref{fig:result_pruning_slot}, we provide the intent accuracy and the slot micro F1 score for the joint intent classification and slot filling tasks as complementary results to sub-section \ref{sec:result_nlu}.

\begin{figure*}[ht!]
\includegraphics[width=0.95\linewidth]{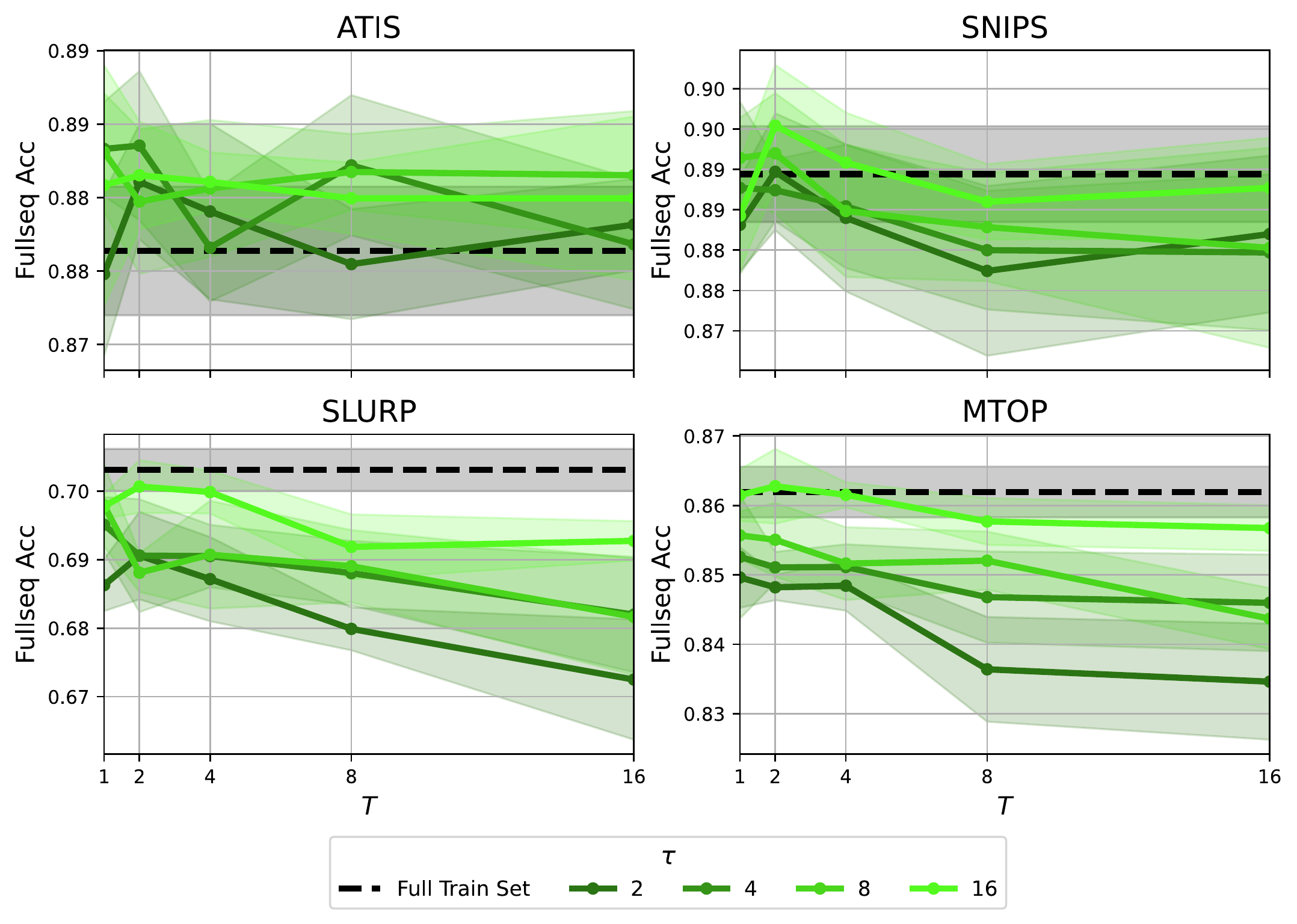}
\caption{Full-sequence accuracy achieved on 40 epochs using our dynamic pruning (EL2N with EMA) across $T$ and $\tau$. We fixed the pruning rate to 70\%.}
\label{fig:result_freq_all}
\end{figure*}

\begin{figure*}[ht!]
\includegraphics[width=0.95\linewidth]{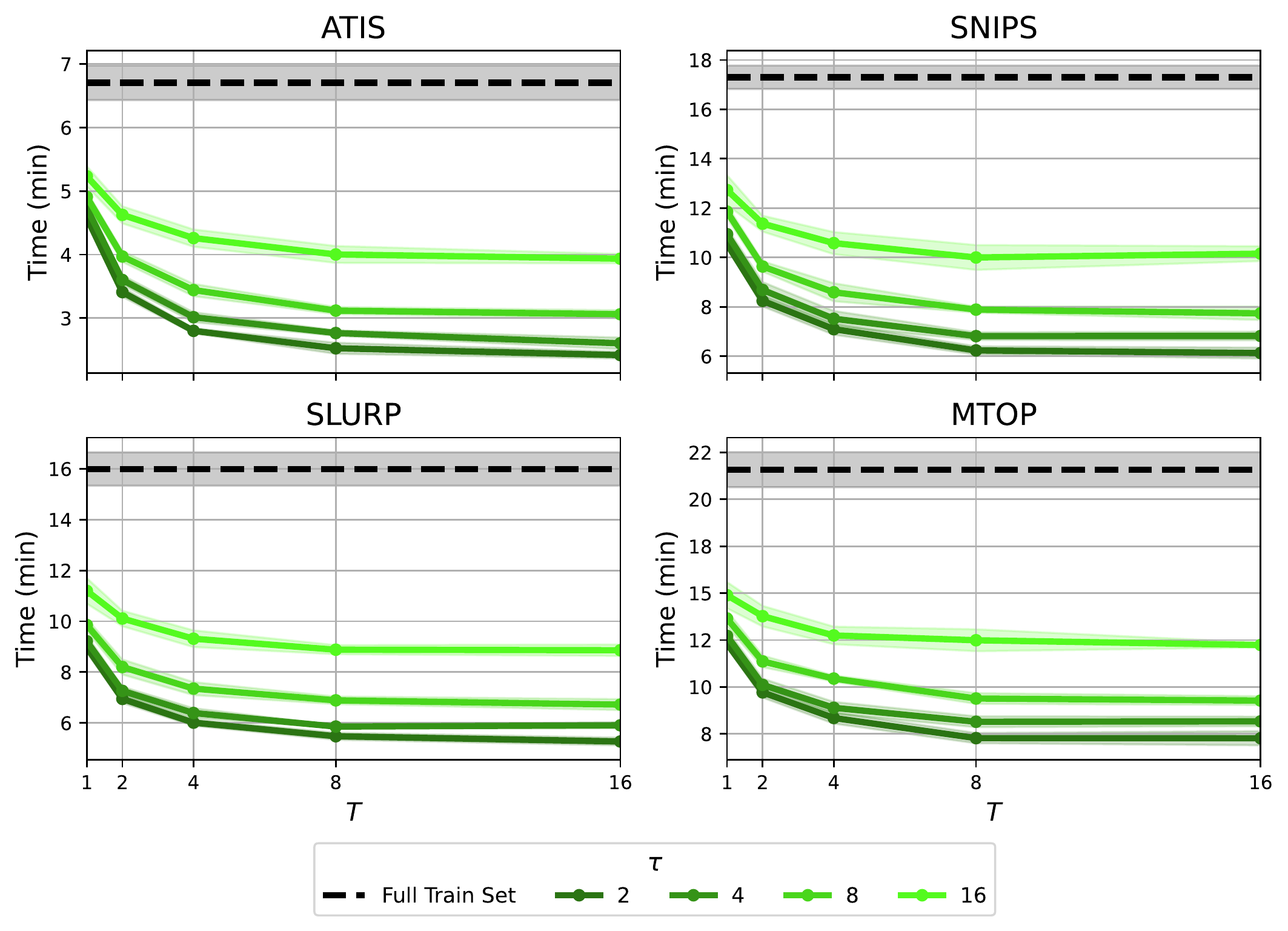}
\caption{Finetuning time achieved on 40 epochs using our dynamic pruning (EL2N with EMA) across $T$ and $\tau$. We fixed the pruning rate to 70\%.}
\label{fig:result_freq_time_all}
\end{figure*}

In Figures \ref{fig:result_freq_all} and \ref{fig:result_freq_time_all}, we present the extensive results on all joint NLU datasets for the experiments on the prune epoch cycle $T$  and prune epoch $\tau$.

In table \ref{tab:datamap}, we present representative examples of hard-to-learn, easy-to-learn and ambiguous samples from fine-tunings on the SLURP dataset, as discussed in section \ref{sec:select_analysis}. 

\begingroup
\setlength{\tabcolsep}{4.5pt} 
\renewcommand{\arraystretch}{2.2} 
\begin{table*}[ht]
\centering
\caption{Examples for hard-to-learn, easy-to-learn and ambiguous samples for the SLURP dataset. \# selection column shows the number of times a sample is selected in retained subset over ten pruning cycles for four different initializations.}
\small
\begin{tabular}{llllll}
\toprule
 & \textbf{Id} & \textbf{Intent label} & \textbf{Annotated utterance} & \textbf{\# selections} & \textbf{\textit{Av.} $\hat{X}_{nlu}$} \\
\midrule
\multirow{4}{*}{\rotatebox[origin=c]{90}{Hard-to-Learn}} 
& 1  & \textit{calendar\_set} & olly play {[}\textit{song\_name} : be warned{]} by {[}\textit{artist\_name} : tech nine{]} & 10, 10, 10, 10 & 1.291                
\\
& 2  & \textit{play\_radio} & play my favorite {[}\textit{device\_type} : radio station{]}  & 10, 10, 10, 10  & 1.018
\\
& 3  & \textit{cooking\_query} & what's the easiest and quickest way to cook a turkey & 10, 10, 10, 10 & 1.255                
\\
& 4  & \textit{general\_joke}  & tell me a joke {[}\textit{joke\_type} : about chickens{]} & 10, 10, 10, 10 & 0.984                
\\ \hline
\multirow{3}{*}{\rotatebox[origin=c]{90}{Easy-to-Learn}} & 5  & \textit{calendar\_set}  & please add this event to my calendar                                     & 0, 0, 0, 0                             & 0.008                \\
                               & 6  & \textit{calendar\_set}  & add an event                                                             & 0, 0, 0, 0                             & 0.012                \\
                               & 7  & \textit{calendar\_set}  & i need you to add this event to my calendar                              & 0, 0, 0, 0                             & 0.08                 \\ \hline
\multirow{12}{*}{\rotatebox[origin=c]{90}{Ambiguous}}

& 8  & \textit{calendar\_set}  & mark {[}\textit{relation} : dad's{]} {[}\textit{event\_name} : retirement dinner{]} for  & 3, 5, 4, 5 & 0.116  \\ & & & {[}\textit{date} : april fourth{]} &   &     
\\
& 9  & \textit{calendar\_set} & add {[}\textit{person} : mary's{]} {[}\textit{event\_name} : birthday{]} on    & 4, 2, 1, 4   & 0.152 \\
& & & the {[}\textit{date} : twenty second{]} to my calendar please  &    &                  
\\ 
& 10  & \textit{calendar\_set}  & can you add my {[}\textit{relation} : brother's{]} {[}\textit{event\_name} : birthday dinner{]} & 5, 3, 6, 6 & 0.224\\
& & & at {[}\textit{place\_name} : rusk{]} for {[}\textit{date} : march twenty third{]} &   &                  
\\
& 11  & \textit{calendar\_set} & remind me about my {[}\textit{date} : monday{]} {[}\textit{event\_name} : meeting{]} with & 6, 9, 5, 8   & 0.299 \\
& & &  {[}\textit{person} : peter francis{]} {[}\textit{time} : fifteen minutes{]} before the meeting  &    &                  
\\ 
& 12  & \textit{calendar\_set}  & add {[}\textit{event\_name} : lee's birthday{]} to the calendar on  & 7, 4, 8, 7 & 0.369 \\
& & &  {[}\textit{date} : twenty two june{]} &   &  
\\
& 13  & \textit{calendar\_set} & please give me a two hour warning before {[}\textit{date} : next saturdays{]} & 8, 8, 9, 8   & 0.530 \\
& & & {[}\textit{event\_name} : meeting{]}  &    &                  \\ 

\bottomrule
\end{tabular}
\label{tab:datamap}
\end{table*}
\endgroup

\end{document}